\documentclass{article}
\usepackage[utf8]{inputenc} 
\usepackage[T1]{fontenc}    
\usepackage{hyperref}       
\usepackage{url}            
\usepackage{booktabs}       
\usepackage{amsfonts}       
\usepackage{nicefrac}       
\usepackage{microtype}      
\usepackage{xcolor}         
\usepackage[pdftex]{graphicx}
\usepackage{amsmath}
\usepackage{amssymb}
\usepackage{mathtools}
\usepackage{amsthm}
\usepackage{subcaption}
\usepackage{multirow}
\usepackage{makecell}
\usepackage{graphics, enumitem}
\usepackage[ruled,vlined,linesnumbered]{algorithm2e} 

\usepackage{caption}
\usepackage{xspace}
\usepackage{bbm}
\usepackage{multibib}

\usepackage[accepted]{mlsys2024}

\SetAlgoNlRelativeSize{0}

\def\ie{{i.e.}}
\def\eg{{e.g.}}

\def\DoC{\mathrm{DoC}}
\def\act{\mathrm{act}}
\def\sim{\mathrm{sim}}

\usepackage{pifont}
\usepackage{multirow}
\usepackage{tikz}
\newcommand*\circled[1]{\tikz[baseline=(char.base)]{
            \node[shape=circle,draw,inner sep=1pt] (char) {#1};}}

\newenvironment{denseitemize}{
\begin{itemize}[topsep=2pt, partopsep=0pt, leftmargin=1.5em]
  \setlength{\itemsep}{2pt}
  \setlength{\parskip}{0pt}
  \setlength{\parsep}{0pt}
}{\end{itemize}}

\def\name{FedTrans\xspace}

\newcommand{\CR}[1]{\textcolor{black}{#1}}
\def\cell{\textit{Cell}\xspace}

\mlsystitlerunning{FedTrans: Efficient Federated Learning Over Heterogeneous Clients via Model Transformation}

\begin{document}

\twocolumn[
\mlsystitle{\name: Efficient Federated Learning via Multi-Model Transformation}

\begin{mlsysauthorlist}
  \mlsysauthor{Yuxuan Zhu}{uiuc}
  \mlsysauthor{Jiachen Liu}{umich}
  \mlsysauthor{Mosharaf Chowdhury}{umich}
  \mlsysauthor{Fan Lai}{uiuc}
\end{mlsysauthorlist}

\mlsysaffiliation{umich}{Computer Science and Engineering,
University of Michigan, Michigan, USA}
\mlsysaffiliation{uiuc}{Computer Science,
University of Illinois at Urbana-Champaign, Illinois, USA}

\mlsyscorrespondingauthor{Yuxuan Zhu}{yxx404@illinois.edu}



\vskip 0.3in

\begin{abstract}
Federated learning (FL) aims to train machine learning (ML) models across potentially millions of edge client devices. 
Yet, training and customizing models for FL clients is notoriously challenging due to the heterogeneity of client data, device capabilities, and the massive scale of clients, making individualized model exploration prohibitively expensive. 
State-of-the-art FL solutions personalize a globally trained model or concurrently train multiple models, but they often incur suboptimal model accuracy and huge training costs. 

In this paper, we introduce \name, a multi-model FL training framework that automatically produces and trains high-accuracy, hardware-compatible models for individual clients at scale.
\name begins with a basic global model, identifies accuracy bottlenecks in model architectures during training, and then employs model transformation to derive new models for heterogeneous clients on the fly. It judiciously assigns models to individual clients while performing soft aggregation on multi-model updates to minimize total training costs. 
Our evaluations using realistic settings show that \name improves individual client model accuracy by \CR{14\% - 72\%} while slashing training costs by \CR{$1.6\times$ - $20\times$} over state-of-the-art solutions.

\end{abstract}

]

\printAffiliationsAndNotice{}

\section{Introduction}
\label{sec:intro} 

Federated learning (FL) is an emerging machine learning (ML) paradigm that trains ML models across potentially millions of clients (\eg, smartphones) over hundreds of training rounds~\cite{flsys1, meta_flsys}. 
In each round, a (logically) centralized coordinator selects a subset of clients and sends the global model to these participating clients (participants). 
Each participant trains the model on its local data and uploads the model update to the coordinator. 
Before advancing to the next round, the coordinator aggregates individual client updates to update the global model. As such, federated learning circumvents systemic privacy risks of cloud-centric ML and high costs of data migration to the cloud~\cite{fedavg, sol}. 

Despite sharing similar goals with traditional cloud ML (\eg, better model accuracy and less resource consumption), FL models are often trained and later deployed on clients with vast device and data heterogeneity~\cite{flint-mlsys, oort, flamingo}. 
This leads to new systems and ML challenges to tailor models for individual clients. 
First, the heterogeneous capabilities of client devices, such as communication and computation, necessitate FL models with different complexities aligned to clients' hardware for better user experience (\eg, model training and inference latency).
Additionally, independently generated data among clients results in heterogeneous data volumes and distributions, making it challenging to train models that fit individual client data at scale.

State-of-the-art FL solutions optimize for better model convergence~\cite{prox}, accuracy fairness~\cite{qfedavg}, and faster round execution~\cite{meta_async}, while often focusing on the performance of a single (global) model.  
This may not suit individual client's device capability and data (\S\ref{sec:background}). 
Although recent work attempted to mitigate data heterogeneity by tuning model weights within client devices~\cite{pfl1, pfl2}, \emph{they overlook client system heterogeneity}, leading to impractically large models for resource-constrained clients and vice versa. 
Recent multi-model approaches train models with different weights~\cite{auxo-socc} or architectures for various clients~\cite{heterofl,splitmix,fedoras}, but they incur high training costs and/or struggle to identify the right model for each client. 

In this paper, we propose \name, a multi-model FL training framework, to automatically and efficiently train customized models for individual clients at scale (\S\ref{sec:overview}). 
\name begins with a small model and progressively expands models in flight to produce well-trained models with different complexities, delivering high-accuracy models for diverse hardware-capable clients at a low cost.

\name addresses the following fundamental requirements toward practical FL deployment (\S\ref{sec:design}). 
First, we must cost-efficiently maximize model accuracy for individual clients subject to their hardware capabilities. 
Unlike existing methods that rely on pre-determined model architectures~\cite{heterofl,splitmix} or prohibitively expensive model exploration~\cite{fednas}, \name identifies accuracy bottlenecks of the training model -- model architecture blocks (\eg, convolutional layers) that are incapable of fitting on clients' data -- using training feedback (\eg, layer gradients). It then adaptively transforms the model architecture, such as by widening or deepening the bottleneck layers, to maximize accuracy improvement for clients with more resources. 
Our transformation customizes for different client groups and effectively warms up new model weights, reducing training costs. 
Second, we must minimize the cost of training multiple models. 
To this end, \name identifies suitable timing for generating new models to maximize the effectiveness of warmup, and performs aggregation of model weights across models to accelerate training convergence, by exploiting model architectural similarity.

We have integrated \name with FedScale\footnote{https://github.com/SymbioticLab/FedScale}~\cite{fedscale} and evaluated it across various FL tasks with real-world workloads (\S\ref{sec:eval}). 
Compared to state-of-the-art FL solutions~\cite{heterofl,splitmix,fluid}, \name improves model accuracy by \CR{14\% - 72\%} and achieves \CR{$1.6\times$ - $20\times$} lower training costs, while reducing manual efforts by automatically spawning models for FL clients at scale.

Overall, we make the following contributions in this paper:

\begin{denseitemize}
    \item We propose a systematic multi-model framework to train performant models for individual clients at scale.
    \item We introduce a novel utility-based model transformation mechanism to generate, train, and assign FL models. 
    \item We evaluate \name in various real-world settings to show its resource savings and accuracy gains.
\end{denseitemize}

\section{Motivation}
\label{sec:background}

We start with an introduction to the challenges in FL (\S\ref{sec:bg-challenge}), followed by the limitations of existing solutions (\S\ref{sec:bg-limitation}).

\subsection{Challenges of Federated Learning}
\label{sec:bg-challenge}

Unlike traditional cloud ML, FL trains and later deploys models on clients with vast heterogeneities in device capabilities and data. This leads to a pressing need for minimizing costs and maximizing model accuracy for individual clients. 

\begin{figure}[t]
    \centering
    \begin{subfigure}{0.23\textwidth}
        \centering
        \includegraphics[width=\linewidth]{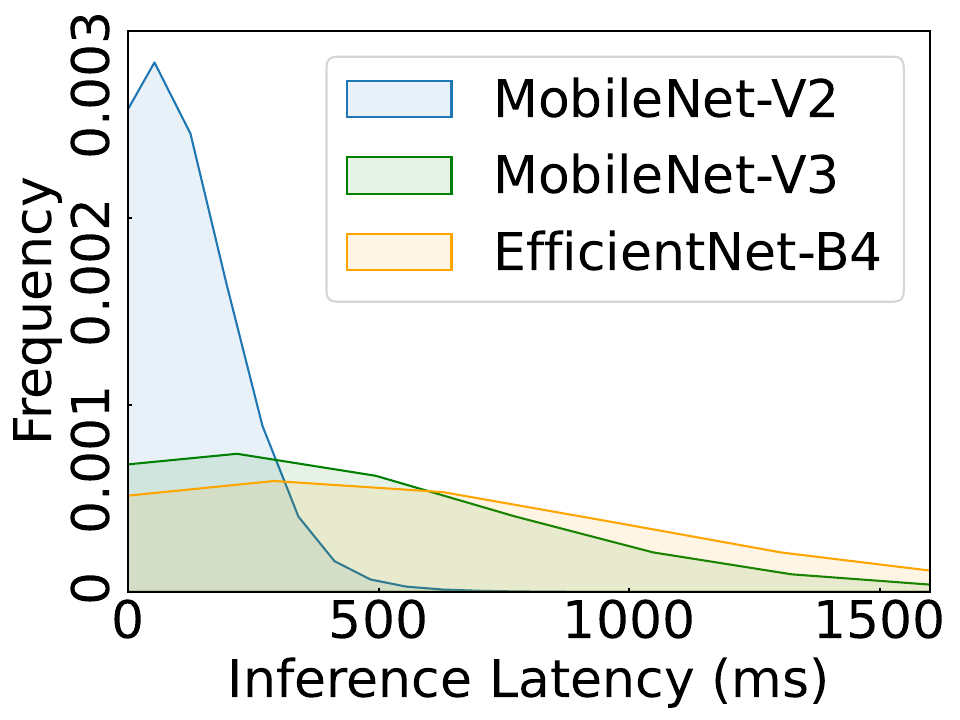}
        \caption{The heterogeneous device capabilities require models with different complexities.}
        \label{fig:task-latency}
    \end{subfigure}%
    \hfill
    \begin{subfigure}{0.23\textwidth}
        \centering
        \includegraphics[width=\linewidth]{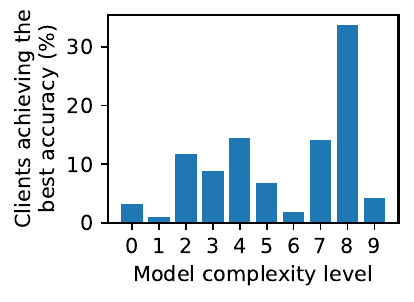}
        \caption{\CR{No single model achieves the best accuracy for the majority of clients.}}
        \label{fig:model-accuracy}
    \end{subfigure}
    \caption{\CR{Client system heterogeneity and accuracy variance call for different models for individual clients.}}
    \label{fig:both-figures}
\end{figure}

\paragraph{System heterogeneity requires models with different complexities.} 
FL client devices often have diverse hardware capabilities, requiring models of different levels of complexity for practical training and deployment.
We analyze the inference latency of three on-device models across over 700 smartphones using the \textit{AI Benchmark}~\cite{AIbench}. 
Figure~\ref{fig:task-latency} shows that end-user devices impose clear model complexity requirements to meet latency constraints. 
For example, users with resource-constrained devices can hardly endure seconds of inference latency from models like MobileNet-V3 in real deployment~\cite{walle-osdi}. 
Hence, different client groups want models of differing complexities. 
Furthermore, clients with the same latency requirements can have multiple architecture choices, as indicated by the distribution overlap (\eg, MobileNet-V3 and EfficientNet-B4). 
Consequently, FL developers must carefully design model architectures for individual clients to maximize accuracy, implying significant human effort.

\paragraph{\CR{Client accuracy variance implies no one-size-fits-all model.}}
The need for training multiple models is further amplified by the client accuracy variance in single-model FL. 
To exemplify this challenge, we experiment with seven models of varying complexities selected from the NASBench201 benchmark~\cite{nasbench201}. 
Here, model complexity is quantified by the number of multiply-accumulate (MACs) operations, with each increase doubling this count. 
We train them on the Federated-EMNIST (FEMNIST) dataset across 3400 clients. 
\CR{Figure \ref{fig:model-accuracy} reports the percentage of clients that achieve the best accuracy on each model.}
\CR{We observe that no single model achieves the best accuracy for the majority of clients.
The model that achieves the highest accuracy varies for different clients, highlighting the need for tailoring models to each client's specific requirements.}
\subsection{Limitations of Existing Solutions}
\label{sec:bg-limitation}

Prior FL work~\cite{prox, qfedavg, meta_async} has primarily focused on improving the performance of a single global model, with recent strides towards multi-model federated learning~\cite{fedoras}. However, existing solutions exhibit shortcomings in FL training costs and model accuracy. 

\begin{figure}
    \centering
    \includegraphics[width=\linewidth]{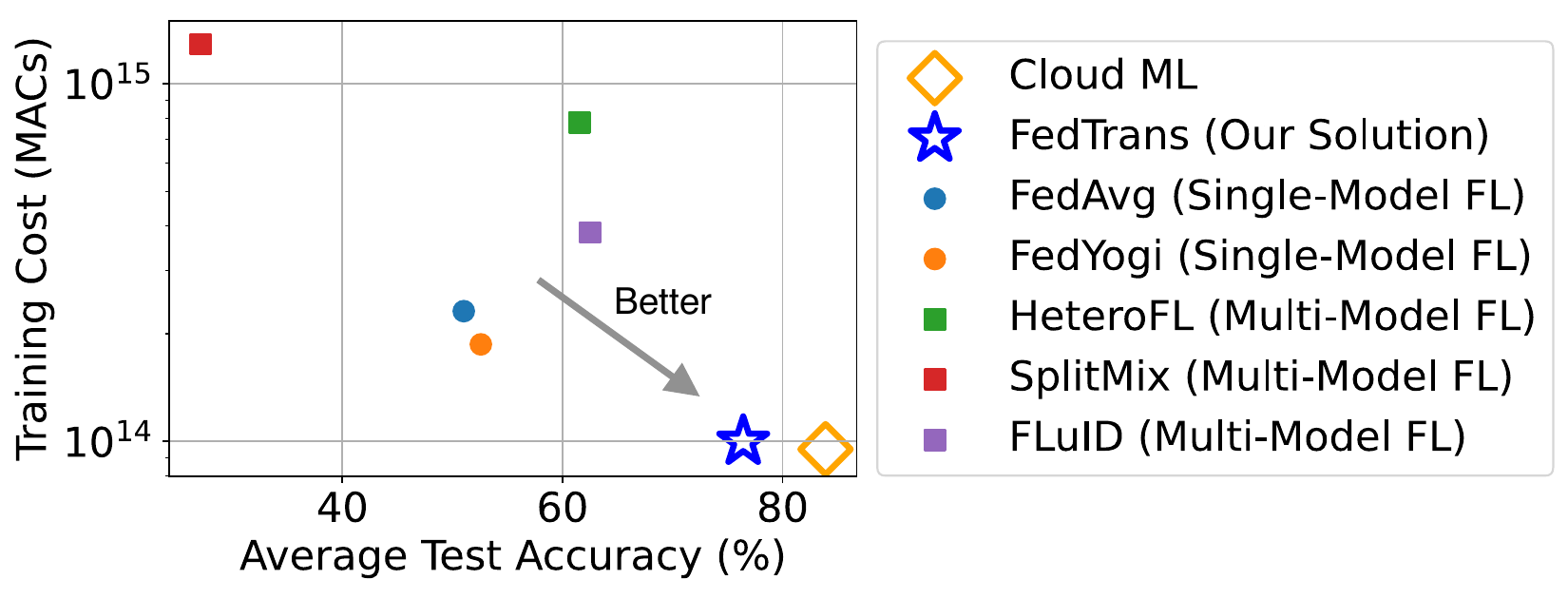}
    \caption{Existing solutions are suboptimal for FL clients.}
    \label{fig:suboptimal}
\end{figure}

\paragraph{Huge training costs and laboring effort.} 
Recent FL advances design and train heterogeneous model architectures for different clients. However, they often require manually determining the model set used for training in advance~\cite{splitmix,fedrolex,fedhm}, a labor-intensive but ineffective process without extensive exploration. 
Moreover, training multiple models concurrently and/or conducting continuous model testing for informed client-model assignment can raise huge costs. Figure~\ref{fig:suboptimal} shows that even state-of-the-art multi-model solutions~\cite{splitmix, heterofl} \CR{and dropout-based solutions~\cite{fluid}} result in a cost increase of orders of magnitude compared to training a single global model. We use multiply-accumulate operations (MACs) to measure costs~\cite{mlsys-mac}.

\paragraph{Inferior model accuracy.} 
Training a single global model may reduce costs, but it struggles to accommodate individual client data (Figure~\ref{fig:suboptimal}). 
Conversely, existing multi-model FL solutions, particularly those using pre-determined models, struggle to pinpoint accuracy bottlenecks in model architectures and/or decide appropriate model assignments to clients at scale.  
When we consider the cloud ML performance as a hypothetical performance upper bound, where the data is centralized and shuffled to be homogeneous, Figure~\ref{fig:suboptimal} shows that existing solutions are far from achieving this upper bound. 

\begin{figure*}[t]
    \centering
    \includegraphics[width=0.9\linewidth]{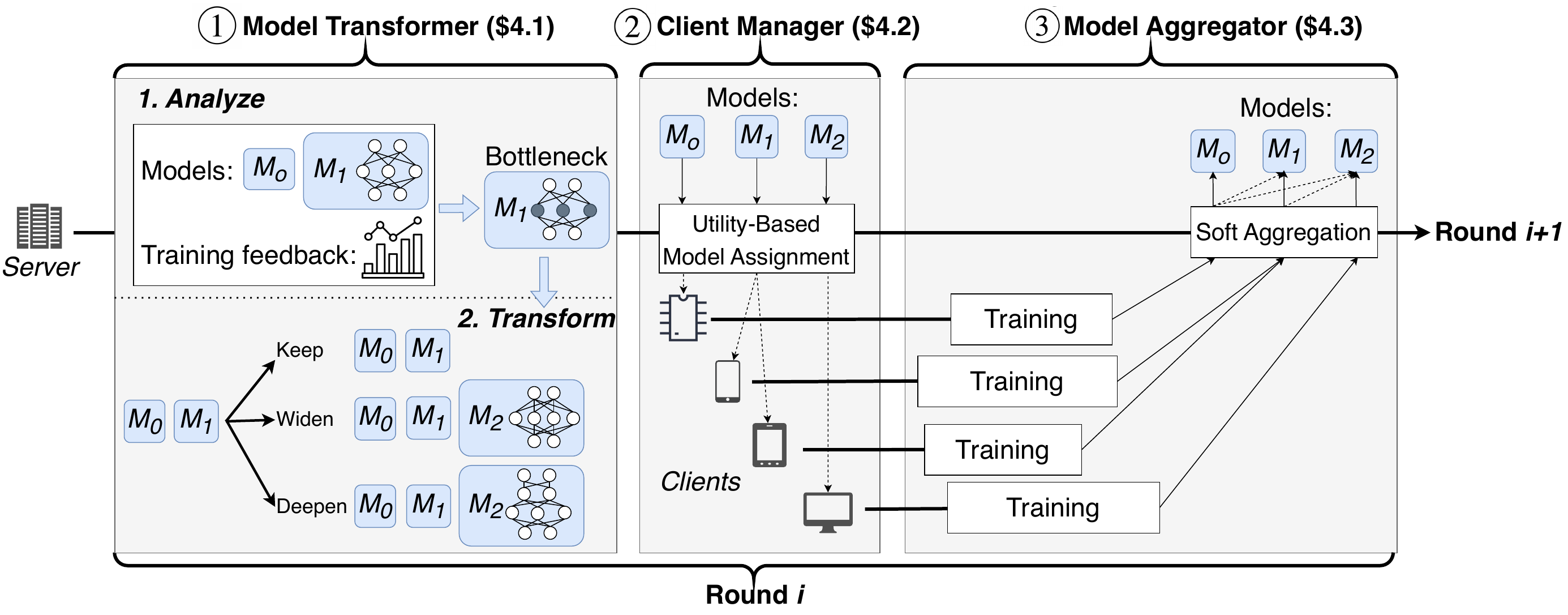}
    \caption{\name architecture and its lifecycle in FL deployment.}
    \label{fig:lifecycle}
\end{figure*}
\section{\name Overview}
\label{sec:overview}
We present \name, a multi-model FL training framework, to efficiently produce and train high-accuracy, hardware-compatible models for individual clients at scale. We next provide an overview of how \name fits in the FL lifecycle. 
\paragraph{System Components}
At its core, \name relies on a model abstraction, \cell, to transform the initial global model into multiple models during FL training. The \cell is the minimum component of the model architecture (\eg, a convolution layer or a ResNet block), on which \name performs model transformation (\eg, widen or deepen a \cell). \name enables online model transformation via three system components:
\begin{denseitemize}
\item 
\emph{Model Transformer}: It captures accuracy bottlenecks in the model architecture, whereby it performs information-preserving transformations on the model architecture to generate new models and warm up model weights.
\item 
\emph{Client Manager}: It explores the utility of each model to individual clients over different training rounds, considering potential accuracy gains and system constraints. Then, it assigns the right model to each client.
\item 
\emph{Model Aggregator}: It manages the training of multiple models and performs soft aggregation on model weights across models to accelerate their convergence. 
\end{denseitemize}
\paragraph{\name Lifecycle}
Figure~\ref{fig:lifecycle} illustrates the lifecycle of \name. FL developers initialize FL training with a small model. 
In each training round, \circled{1} 
Model Transformer leverages training feedback from previous rounds (\eg, gradient and loss) to potentially transform (\eg, widen or deepen) a specific \cell of the model. 
If transformation occurs, Model Transformer will generate a new model with the current largest model weights to reduce training costs. 
\circled{2}
For each participant, Client Manager assigns an appropriate model among all currently available models, in terms of their utility (\eg, training loss) and system constraints. Then participants download their models and start local training. 
\circled{3}
When receiving client updates, Model Aggregator aggregates model weights in a weight-sharing manner to accelerate convergence. 

FedTrans iterates these phases in each round until exhausting the training budget, or the model architecture complexity reaches the maximum supported by any participant and all models converge. 
Note that conventional single model training is a special case of this lifecycle, where \name does not create any new model. 
\section{\name Design}
\label{sec:design}

In this section, we introduce how Model Transformer generates a suite of performant models at low training costs (\S\ref{sec:transformer}), how Client Manager assigns individual clients with the right model (\S\ref{sec:client-manager}), and how Model Aggregator accelerates the co-training of model suites (\S\ref{sec:aggregator}).

\subsection{Model Transformer}
\label{sec:transformer}

Generating the right model for FL clients is challenging due to the interplay among model accuracy, client system constraints, and training costs. Intuitively, using a large model often implies better accuracy, but it incurs high training costs, and clients may not be able to run such large models~\cite{flint-mlsys}. It is tempting to model it as a Neural Architecture Search (NAS) problem in traditional cloud ML, such as by training a super-network~\cite{google-darts}. However, doing so in FL is suboptimal and even impractical due to the expensive planet-scale FL training (\eg, training a super-network) and massive data heterogeneity. 

To strike a balance between low costs and high model accuracy, Model Transformer generates new models with different system requirements by transforming, thus improving model architectures during FL training. 
The key insight is that (1) models achieving state-of-the-art performance are often built with well-established blocks (\eg, ResNet or Transformer blocks), so we can reduce the exploration space of model architectures by transforming at the block (\cell) level; and (2) model transformation, which changes the width and depth of bottleneck {\cell}s to generate a new model with inherited model weights~\cite{modelkeeper}, can reduce training costs by warming up new models' training. 

However, we need to tackle the following challenges: 
\begin{denseitemize}
\item When to perform a model transformation? 
\item At which \cell of the model architecture to transform? 
\item How to transform (\eg, widen/deepen) the selected \cell?
\end{denseitemize}

\paragraph{Identifying the right time to transform}
When to transform introduces a trade-off between the maturity of the current model and the waiting time spent on training new models. 
Transforming the current model too early yields limited warmup benefits due to the current model's low accuracy, resulting in longer training of the new model. 
On the other hand, transforming too late, such as when model training converges, results in longer waiting times to start the new model and only marginal warm-up benefits due to the sublinear convergence speed of ML training.

To find the sweet spot, we refer to the popular elbow method that picks the elbow of the curve to maximize benefits~\cite{elbow-clustering}. 
Here, we define the degree of convergence ($\DoC$) to measure the slope of the moving training loss. Equation \ref{eq:trans-cri} defines the $\DoC$ at round $i$ by averaging $\gamma$ consecutive loss ($L$) slopes, where each loss slope is calculated with a step of $\delta$: 
\begin{equation}
    \DoC = \frac{1}{\gamma}\sum_{i=1}^\gamma \frac{L(i-\delta) - L(i)}{\delta}
    \label{eq:trans-cri}
\end{equation}
\name initiates the transformation when $\DoC$ is less than a certain threshold, \ie, reaching the elbow of the curve. \CR{Intuitively, a larger threshold will make \name transform models more frequently, while a larger step size tolerates more oscillations of the training loss.} Theoretically, our DoC design draws from the positive link between loss sharpness and model generalizability \cite{visualizingloss, lossbounds}. A smooth loss curve indicates robust generalizability and guides optimal model transformation timing. We also empirically evaluate the effectiveness of our $\DoC$ design (\S\ref{subsec:ablation}).

\paragraph{Picking the right model \cell to transform}
Once starting the transformation, we need to decide to transform which model \cell can unblock model accuracy with the limited information in practical deployment, where only the training loss and gradients are available most of the time. In complement to existing privacy-preserving mechanisms, \name solely utilizes aggregate gradients, not the gradients of individual clients.

Model Transformer selects the \textit{Cells} with larger gradient norms to transform. Intuitively, a large gradient implies that \cell is under great dynamics and harder to fit, especially given that transformation will not be activated unless the model starts to converge, so we should transform this \cell to unblock model convergence (performance). 
In theory, our design is motivated by transfer learning, where layers that are actively being updated during training contribute more to the bottleneck of model convergence~\cite{transfer1, transfer2}, indicating the incapability of capturing the data characteristics of certain FL clients. 

\begin{figure}[t]
    \centering
    \includegraphics[width=0.9\linewidth]{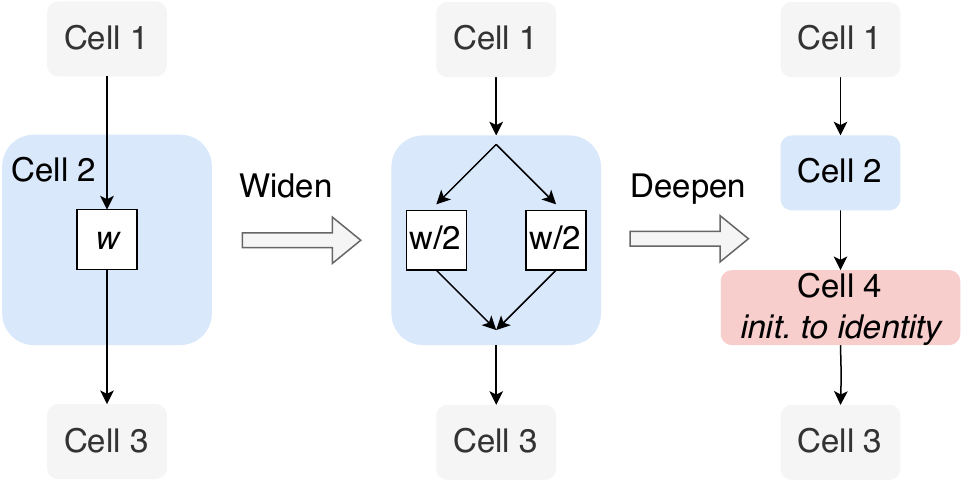}
    \caption{Model transformation on a convolution \cell.}
    \label{fig:trans_example}
\end{figure}

\name relies on the distribution of \cell activeness -- measured by the gradient norm $\frac{\|\nabla w_l\|}{\|w_l\|}$ in round update -- to decide which \textit{Cells} to transform. We normalize the gradient norm by the weight norm to mitigate the bias in selecting cells due to gradient vanishing. Since a model can have multiple \textit{Cells} blocking the performance, we transform the \textit{Cells} whose activeness exceeds $\alpha$ times of the largest activeness among all \textit{Cells}. By default, $\alpha$ is set to 0.9.
We empirically evaluate the effectiveness of $\alpha$ in our ablation study (\S\ref{subsec:ablation}).
%
\paragraph{Widening or Deepening}
After selecting which \textit{Cells} to transform, we may widen or deepen them. 
Taking Figure \ref{fig:trans_example} as an example, we can widen a \cell by increasing the number of neurons or deepen it by inserting an identity \cell, which directly passes the input of its predecessor to the successor. 
However, doing so is non-trivial. First, we need to decide which operation to perform and enable it with little overhead and loss of its parent model's weights. Also, we should carefully control the degree of transformation to avoid growing too fast and missing the right complexity for clients. 

\begin{figure}[t]
    \centering
    \includegraphics[width=0.9\linewidth]{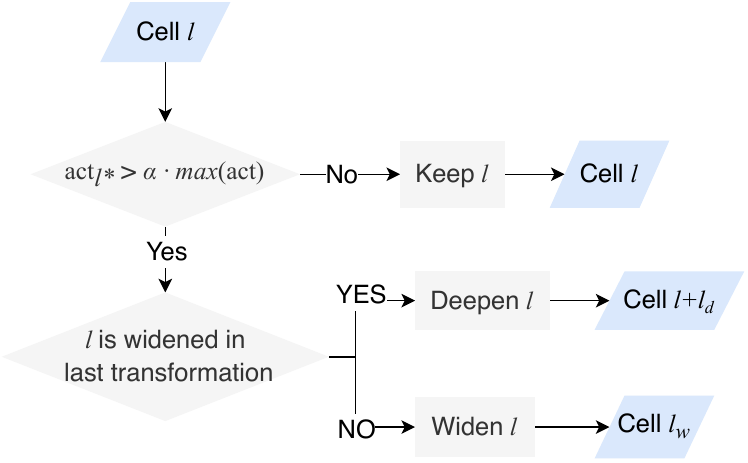}
    \caption{Control flow of the transformation for a \cell $l$.}
    \label{fig:trans}
\end{figure}

Inspired by the concept of compound scaling in model architecture design~\cite{efficientnet}, which emphasizes the importance of balanced width and depth for efficiency, Model Transformer alternates between widening and deepening the \cell. 
Figure \ref{fig:trans} shows the overall procedure to perform architecture transformation, wherein each operation will widen the selected \cell by a factor or insert (deepen) a certain number of \cell(s). By default, \name widens a \cell by two or inserts one \cell at a time. After these transformations, \name generates a new training model with the potential for better accuracy upon convergence.  We empirically show that our compound scaling design achieves better performance than its counterparts (\S\ref{subsec:ablation}). 

After widening or deepening the \cell, Model Transformer performs function-preserving weight transformation to inherit model weights. 
When deepening a \cell, we initialize the newly inserted cell's weights as an identity tensor. When widening a \cell, we randomly select columns from the pre-expanded \cell's weights to fill the widened \cell and divide each column by the times it is selected.
This transformation, theoretically~\cite{net2net, ofa, modelkeeper}, maintains the parent model's information, ensuring the same tensor output for many common tensor operators such as fully connected and convolutional layers. As a result, it allows for the transfer of model weights to warm up new models. 

\subsection{Client Manager}
\label{sec:client-manager}
\begin{footnotesize}
\begin{algorithm}[t]
    \DontPrintSemicolon
    \SetKwInOut{Param}{Parameter}
    \SetKwInOut{Input}{Input}\SetKwInOut{Output}{Output}
    \SetKwComment{SideComment}{\textcolor{blue}{//~}}{}
    \SetKwComment{CommentStyle}{\textit{// }}{}
    \SetKwComment{LineComment}{\textcolor{blue}{\textit{/* }}}{\textcolor{blue}{  */}}
    \Param{\cell selection threshold $\alpha$, $\DoC$ threshold $\beta$}
    \Input{initial model with $k_0$ cells $M_0=[l_1^0, \ldots, l_{k_0}^0]$,
           initial model weights $W_0=[w_1^0, \ldots, w_{k_0}^0]$,
           registered client list $\mathcal{C}$,
           client capacity list $T$,
           number of clients per round $N$}
    \Output{a list of models trained $\mathcal{M}$}
    \BlankLine
    $\mathcal{M} \leftarrow [M_0]$, 
    $\mathcal{W} \leftarrow [W_0]$ \hfill  \textcolor{blue}{// init. models \& weights}\; 
    $U_0 \leftarrow [0, \ldots, 0], \mathcal{U} \leftarrow [U_0]$ \textcolor{blue}{// init. utilities}\;    
    \For{each FL training round $i=0, 1, 2, \ldots$}{
        $\widetilde{\mathcal{C}} \leftarrow \mathrm{Select}(\mathcal{C}, N)$\;
        \vspace{5pt}
        \textcolor{blue}{// Assigning models to clients (\S\ref{sec:client-manager})}\label{alg:client}\;
        \For{each selected client $c \in \widetilde{\mathcal{C}}$ \textbf{parallelly}}{ 
            $K \leftarrow \|\{M \in \mathcal{M} | \mathrm{MAC}(M) \le T_{c}\}\|$\;
            $n \leftarrow \mathrm{Sample}(\mathcal{U}_c, [0, \ldots, K-1])$\;
            \textcolor{blue}{// Clients return weights, gradients, and loss}\;
            $W_{n,c}, G_{n,c}, L_{n,c} \leftarrow \mathrm{ClientTrain}(\mathcal{W}[n], c)$\; 
        }
    $\mathcal{U} \leftarrow \mathrm{UpdateUtility}(\mathcal{U}, L)$ \hfill  \textcolor{blue}{// Eq.~(\ref{eq:update})}\;
    \vspace{5pt}
    \textcolor{blue}{// Inter-model weights aggregation (\S\ref{sec:aggregator})}\label{alg:aggregator}\;
    $\mathcal{W} \leftarrow \mathrm{UpdateWeight}(W)$\;
    \vspace{5pt}
    \textcolor{blue}{// Model transformation and warmup (\S\ref{sec:transformer})}\label{alg:transformer}\;
    $\DoC \leftarrow \mathrm{UpdateDoC}(L)$ \hfill  \textcolor{blue}{// Eq.~(\ref{eq:trans-cri})}\;
    \If{$\DoC \le \beta$}{
        $M^* \leftarrow$ copy the parent model's weights\;
        $U^* \leftarrow$ copy the parent model's utility\;
        \For{each \cell $l^* \in M^*$}{
            $\act_{l^*} \leftarrow \mathrm{UpdateAct}(G, W)$ \;
            $\mathrm{TransformCell}(l^*,\act_{l^*}, \alpha)$ \;
        }
        add new model $M^*$ to $\mathcal{M}$\;
    }
    }
    \caption{Pseudo-code of \name runtime.}
    \label{alg}
\end{algorithm}
\end{footnotesize} 
We have discussed how to generate models \emph{in the cloud}, to account for clients with heterogeneous data and system preferences. The following major challenge is to distribute and train these models for individual clients \emph{at the edge}. 
Algorithm~\ref{alg} outlines how \name enables this model transformation across numerous clients through the coordination of Client Manager (Line \ref{alg:client}) and Model Aggregator (Line \ref{alg:aggregator}). 
Next, we introduce how the Client Manager selects the right model among all generated models for each client. 

\paragraph{Utility-Based Model Assignment}
The design of the model assignment should follow two criteria. First, each client should be assigned a model whose complexity does not exceed the client's hardware capability to respect their system constraints. Here, we consider the models whose number of multiply-accumulate operations (MACs) is fewer than the participant's hardware capability, called compatible models. Note that our solution can easily support other definitions of ``compatible models.'' Second, among compatible models, we should assign the model that is most suitable to that client's data characteristics. Moreover, the design should only leverage the information available in today's FL deployment to respect privacy. For example, we cannot access the client data. 

Here, we leverage the training loss of models to approximate their data-aware affinity to individual clients. Intuitively, a small training loss means potentially better accuracy of that model on client data. However, to avoid prematurely overfitting to a specific model, we maintain a loss-based utility list for each registered participant $c$ with $K$ compatible models: $U^c = [U^c_0, \ldots, U^c_{K-1}]$. When the client participates in FL training, the coordinator probabilistically samples a model following the distribution of model loss, as specified in Eq. \ref{eq:sample}. Here, we take the exponential format of training loss for its smoothness and normalize the probability by the sum of all models' training loss on that client (Eq. \ref{eq:prob}): 
\begin{gather}
    n = \mathrm{Sample}(p^c, [0, \ldots, K-1]) \label{eq:sample} \\
    p^c_k = \frac{\exp{[U^c_k]}}{\sum_{k=0}^{K-1} \exp{[U^c_k]}}, k=0, \ldots, K-1 \label{eq:prob} 
\end{gather}

The utilities of each client on each model are updated using the model performance (e.g. training loss). This soft model assignment scheme encourages a client to explore other models when its training performance is bad (\ie, high training loss), while encouraging a client to stick to the current model when its training performance is good.

\paragraph{Joint Utility Learning} 
However, new models can be generated over time, and not all models are trained on that client all the time. Updating the utility $U^c_k$ of each model can be sporadic, and the previous utility can become stale. 
To accelerate the exploration of better model assignment, the Client Manager jointly updates the utility of compatible models based on their architectural similarity. This is because similar models tend to exhibit similar model accuracy. As such, after the completion of that client, the Client Manager updates the utility lists of all the compatible models as follows: 
\begin{equation}
    {U^c_k} = U^c_k - L^c_k \cdot \sim(M_k,M^*)
    \label{eq:update}
\end{equation}
where $M_k$ is the k-th compatible model, $M^*$ is the model assigned to client $c$ in the last round, $L^c_k$ is the standardized training loss of client $c$ in the last round, and $sim(\cdot, \cdot) \in [0,1]$ calculates the similarity between two models. 
We take the subtractive format as a high training loss means a lower utility. 
As such, the model with similar architectures would borrow more utility information from the current model. 

Here, we measure the similarity of two model architectures in terms of the \cell-wise number of parameters that we can transform, which aligns with our design of model transformation. 
For each \cell $l$, we measure its similarity score, $mc(l)$, between the new model and the model it transformed from (\ie, parent model). 
Specifically, (a) if $l$ is inherited from $M$ without change, $mc=1$ (\ie a full matching degree); (b) if $l$ is widened from the \cell $l'$ of $M$, $mc=\#\mathrm{param}(l')/\#\mathrm{param}(l)$ (\ie, the portion of inherited model weights); (c) if $l$ is inserted to $M$ in the deepen operation, then $mc=0$ as it does not inherit any model weights; Otherwise, (d) $mc=-1$ because it loses the weights of its parent model. Finally, we get the model similarity, $\sim(M_k,M^*)$, by cumulating the similarity of all \emph{Cells}. 


\subsection{Model Aggregator}
\label{sec:aggregator}

After assigning models to clients, we concurrently train (co-train) multiple models.  Minimizing their total training costs becomes the crux. 
Moreover, compared to conventional single-model training, each model trains on fewer clients due to the unbalanced model assignment and system compatibility, which can slow down training convergence.

\begin{table}[htbp]
    \centering
    \begin{tabular}{l|c|c}
        \toprule
        Breakdown & Dataset & Avg. Accu. (\%) \\
        \midrule
        \name & FEMNIST & 75.5  \\
        \name (l2s) & FEMNIST & 60.9 \\
        \name & Cifar10 & 77.5 \\
        \name (l2s) & Cifar10 & 54.1 \\
        \bottomrule
    \end{tabular}
    \captionsetup{width=0.9\linewidth}
    \captionof{table}{Model accuracy with or without weights sharing from large models to small models}
    \label{tab:sharing}
\end{table}

Historically, sharing weights among similar model architectures has accelerated model convergence~\cite{heterofl, hetero_nips22, nas_weight_sharing}. 
While this implies an opportunity to leverage the similarity of models, aggregating multi-model weights is non-trivial. This is because not all models contribute equally to individual model aggregation. 
First, \name generates large models when the small model is converging, meaning that the small model is under fine-tuning while the large model is still under-trained. 
Using large models to update small models can lead to big noise in the convergence of small models, impeding their convergence. 
As shown in Table \ref{tab:sharing}, we compare the accuracy performance of FedTrans on  FEMNIST and Cifar10 datasets, with and without enabling weights sharing from large models to small models (l2s). Disabling l2s significantly improves final model accuracy.

Second, when leveraging the weights of different models to update a model, we must consider their architectural similarity and real-time convergence. 

By taking these two insights into account, we develop a soft model aggregation mechanism that performs a \emph{weighted} average over models' weights while considering model architectural similarity:
\begin{equation}
    w_j = \frac{\sum_{i=1}^{j} \eta^{\mathbbm{1}(i\ne j)\times t} \mathrm{sim}(M_i, M_j) \cdot w_i }{\sum_{i=1}^{j} \mathrm{sim}(M_i, M_j)}
    \label{eq:weight_sharing}
\end{equation}
$\eta^{\mathbbm{1}(k\ne j)\times t}$ is the decaying factor in round $t$, while $w_j$ and $w_i$ denote the weights of model $M_j$ and $M_i$, respectively. We crop $w_i$ if necessary to fit the shape of $w_j$ as in HeteroFL~\cite{heterofl}.

When updating model weights using Eq.~(\ref{eq:weight_sharing}) in each round, it combines updates from both that model's clients and similar models based on their architectural similarity. Moreover, as the model converges over rounds, $\eta$ progressively reduces the impact of other models to mitigate noise toward its training convergence. 
We empirically show the effectiveness of our soft aggregation design using realistic datasets (\S\ref{subsec:breakdown}).

\begin{scriptsize}
    \begin{table*}[htbp]
        \centering
        \begin{tabular}{c|crlrrlrr}
            \toprule
            & Method   & Accu. (\%) & & IQR (\%) & Cost (PMACs) & & Storage (MB) & Network (MB)\\ 
            \midrule
            \multirow{4}{*}{\rotatebox{90}{\footnotesize{\textit{CIFAR-10}}}} 
            & FedTrans & 78.29 &                                    & 5.25 & 0.86 &                                          & 6.8 & 1368.4\\
            & FLuID    & 59.81 & \textcolor{blue}{$\uparrow$ 18.48} & 3.00 & 1.50 & \textcolor{blue}{$\downarrow 1.7\times$} & 71.7 & 1433.4\\
            & HeteroFL & 64.51 & \textcolor{blue}{$\uparrow$ 13.78} & 7.00 & 4.31 & \textcolor{blue}{$\downarrow 5.0\times$} & 45.7 & 1828.4\\
            & SplitMix & 51.36 & \textcolor{blue}{$\uparrow$ 26.93} & 6.23 & 4.49 & \textcolor{blue}{$\downarrow 5.2\times$} & 35.4 & 7089.3 \\
            \midrule
            \multirow{4}{*}{\rotatebox{90}{\footnotesize{\textit{FEMNIST}}}} 
            & FedTrans & 76.42 &                                    & 13.23 & 0.10 &                                          & 0.04 & 8.9\\
            & FLuID    & 62.52 & \textcolor{blue}{$\uparrow$ 13.90} & 13.63 & 0.38 & \textcolor{blue}{$\downarrow 3.8\times$} & 0.3 & 64.6\\
            & HeteroFL & 61.54 & \textcolor{blue}{$\uparrow$ 14.88} & 13.50 & 0.78 & \textcolor{blue}{$\downarrow 7.8\times$} & 1.0 & 209.0\\
            & SplitMix & 27.13 & \textcolor{blue}{$\uparrow$ 49.29} & 10.60 & 1.29 & \textcolor{blue}{$\downarrow 12.9\times$} & 1.3 & 9392.5 \\
            \midrule
            \multirow{4}{*}{\rotatebox{90}{\footnotesize{\textit{Speech}}}} 
            & FedTrans & 91.75 &                                    & 4.75  & 0.10 &                                           & 0.3 & 56.4\\
            & FLuID    & 76.50 & \textcolor{blue}{$\uparrow$ 15.25} & 43.65 & 0.26 & \textcolor{blue}{$\downarrow 2.6 \times$} & 0.6 & 121.5\\
            & HeteroFL & 73.69 & \textcolor{blue}{$\uparrow$ 18.06} & 40.00 & 1.49 & \textcolor{blue}{$\downarrow 14.9 \times$} & 1.0 & 197.5\\
            & SplitMix & 19.60 & \textcolor{blue}{$\uparrow$ 72.15} & 8.89  & 0.60 & \textcolor{blue}{$\downarrow 6.0 \times$} & 1.3 & 5736.8\\
            \midrule
            \multirow{4}{*}{\rotatebox{90}{\footnotesize{\textit{OpenImage}}}} 
            & FedTrans & 61.86 &                                    & 28.56 & 47.39  &                                           & 10.6 & 2118.9 \\
            & FLuID    & 32.87 & \textcolor{blue}{$\uparrow$ 28.99} & 51.61 & 76.88  & \textcolor{blue}{$\downarrow 1.6\times$}  & 35.5 & 4926.92 \\
            & HeteroFL & 24.53 & \textcolor{blue}{$\uparrow$ 37.33} & 36.00 & 799.52 & \textcolor{blue}{$\downarrow 16.9\times$} & 92.0 & 6187.7 \\
            & SplitMix & 35.44 & \textcolor{blue}{$\uparrow$ 26.24} & 21.15 & 950.38 & \textcolor{blue}{$\downarrow 20.1\times$} & 38.7 & 2468.0 \\
            \bottomrule    
        \end{tabular}
        \caption{Detailed performance comparison.}
        \label{tab:perf}
    \end{table*}
\end{scriptsize}

\section{Evaluation}
\label{sec:eval}

We evaluate \name on four CV and NLP datasets across four models. We summarize the results as follows:
\begin{denseitemize}
    \item 
    \name improves average model accuracy by \CR{13.78\% - 72.15\%} against existing multi-model FL frameworks, while reducing training costs by \CR{$1.6\times$ - $20.0\times$} (\S\ref{subsec:e2e}).

    \item 
    \name reduces manual efforts by automatically finding better models for clients, wherein each component is effective for the overall performance (\S\ref{subsec:breakdown}). 

    \item 
    \name improves performance over a wide range of settings and outperforms its design counterparts (\S\ref{subsec:ablation}).
\end{denseitemize}

\subsection{Methodology}

\paragraph{Experimental Setup} 
We conducted experiments using FedScale~\cite{fedscale}, a widely used FL benchmarking platform, on a cluster of 15 NVIDIA V100 GPUs. The platform produces realistic FL client system speed and client data. 
For each training round, we select 100 clients, each using a batch size of 10 and performing 20 local steps. More details are available in Appendix~\ref{subsec:expset}. 
The initial model's complexity corresponds to the client with the lowest computation and communication capacities, while the maximum model's complexity aligns with the client possessing the highest resource capacities. \CR{We sample client hardware capacities from FedScale, which includes the traces of 500k real-world mobile devices. The disparity between the most capable and least capable devices exceeds 29$\times$.}

\paragraph{Datasets and Models} 
We first experiment with \emph{CIFAR-10} image classification tasks and follow existing works~\cite{heterofl} to partition them into non-IID datasets with 100 clients. \CR{Then, we run three real-world FL datasets of different scales and use their realistic partitions, which are widely used in FL benchmarking \cite{auxo-socc,fedscale,prox}:}
\begin{denseitemize}
\item
\emph{Speech Command}: a small-scale Google speech dataset of 2,618 clients~\cite{google-speech}. We use ResNet18 as the initial model to recognize 35-category commands. 
\item
\emph{F-EMNIST}: a middle-scale image classification dataset of 3,400 clients~\cite{femnist}. We choose the smallest model in NAS- Bench201 as the initial model.
\item
\emph{OpenImage}: a large-scale image classification dataset of 14,477 clients~\cite{openimg}, with 1.5 million images spanning 600 categories. We start with ResNet18. 
\end{denseitemize}
More detailed setups are available in Appendix~\S\ref{subsec:expset}.
\begin{figure*}[t]
    \centering
    \begin{minipage}{.24\linewidth}
        \centering
        \includegraphics[width=\linewidth]{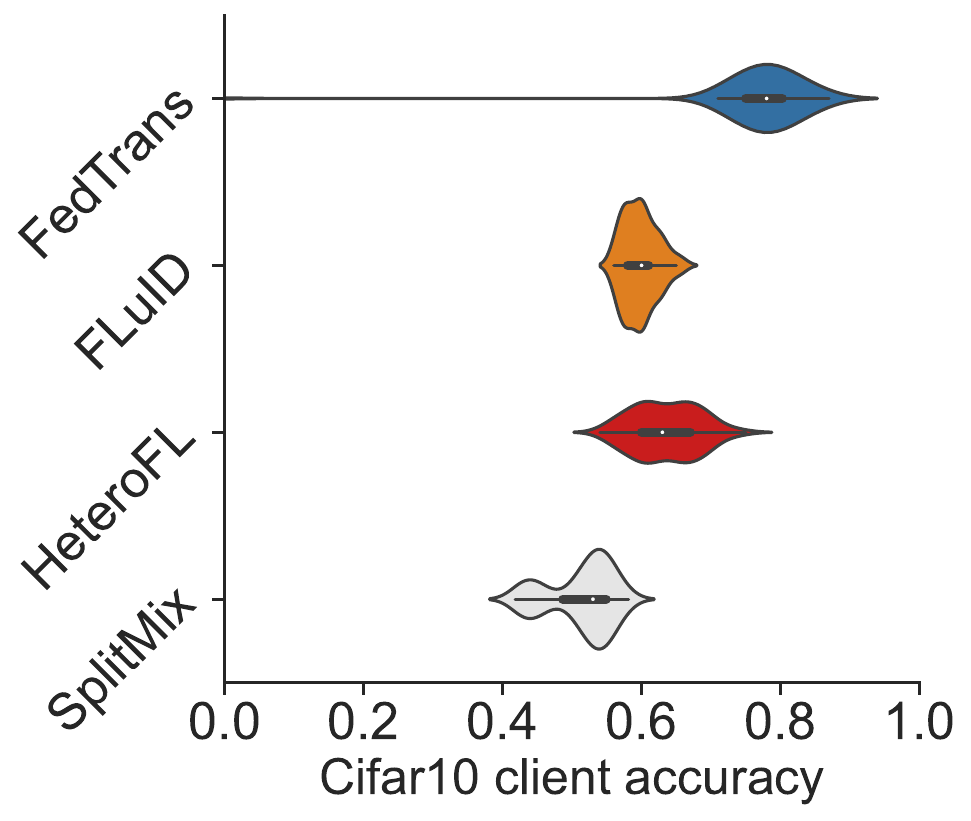}
        \captionsetup{width=0.9\linewidth}
        \label{fig:cifar}
    \end{minipage}%
    \hfill
    \begin{minipage}{.24\linewidth}
        \centering
        \includegraphics[width=\linewidth]{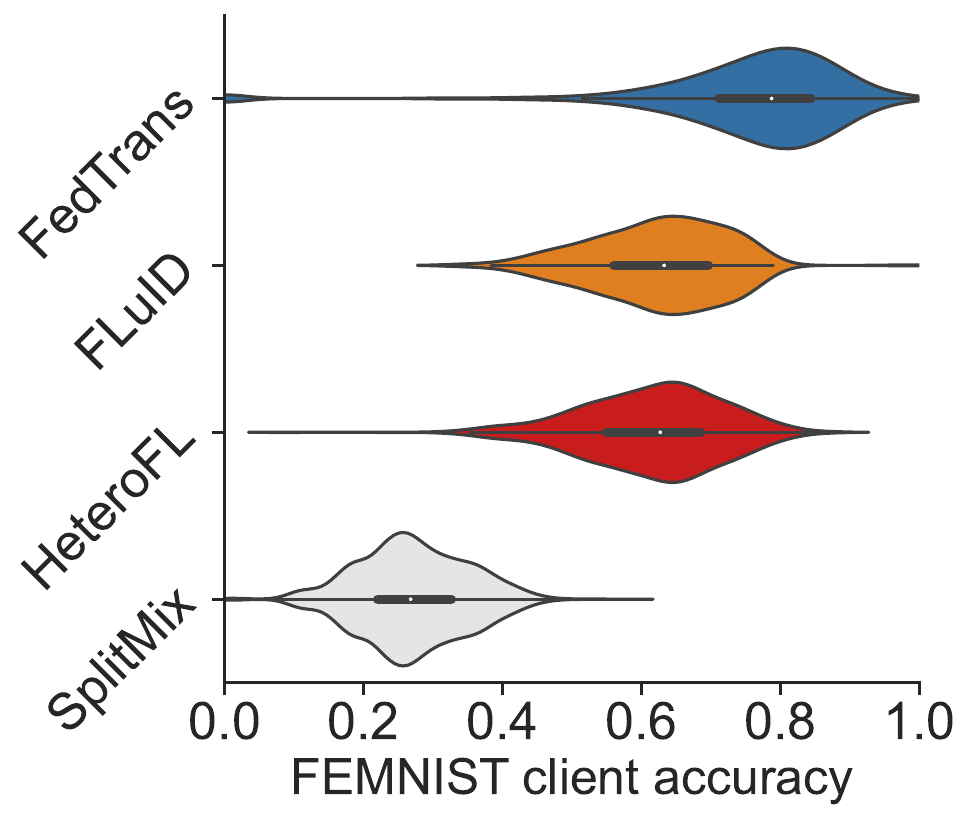}
        \captionsetup{width=0.9\linewidth}
        \label{fig:femnist}
    \end{minipage}%
    \hfill
    \begin{minipage}{.24\linewidth}
        \centering
        \includegraphics[width=\linewidth]{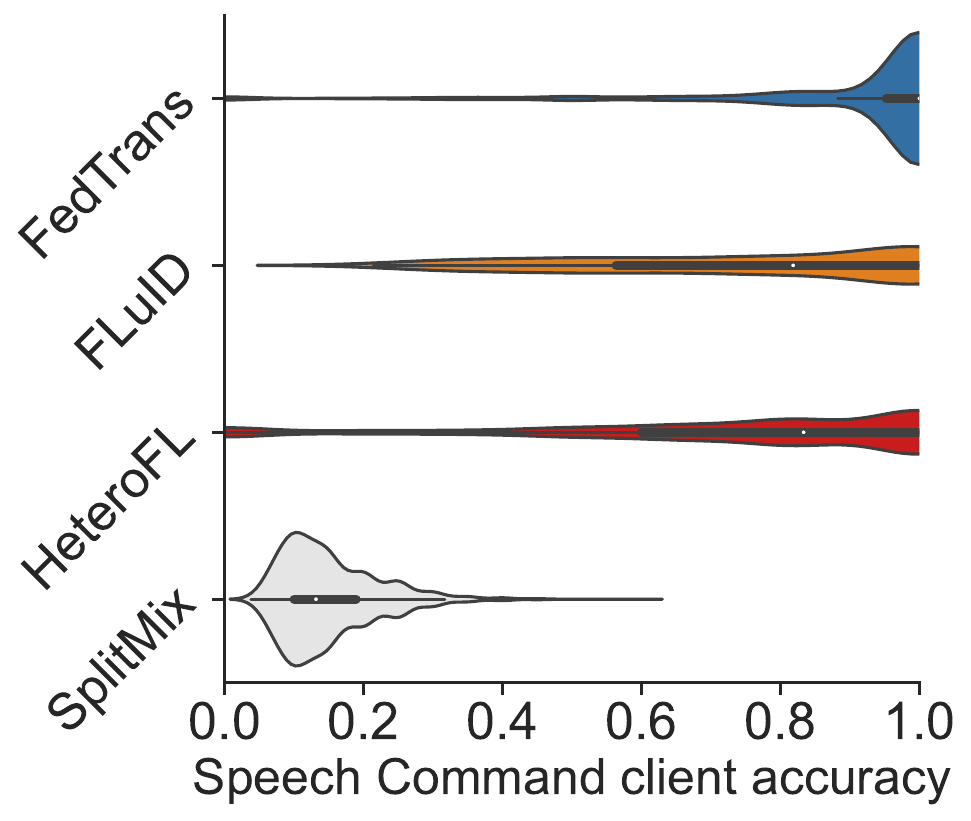}
        \captionsetup{width=0.9\linewidth}
        \label{fig:speech}
    \end{minipage}%
    \hfill
    \begin{minipage}{.24\linewidth}
        \centering
        \includegraphics[width=\linewidth]{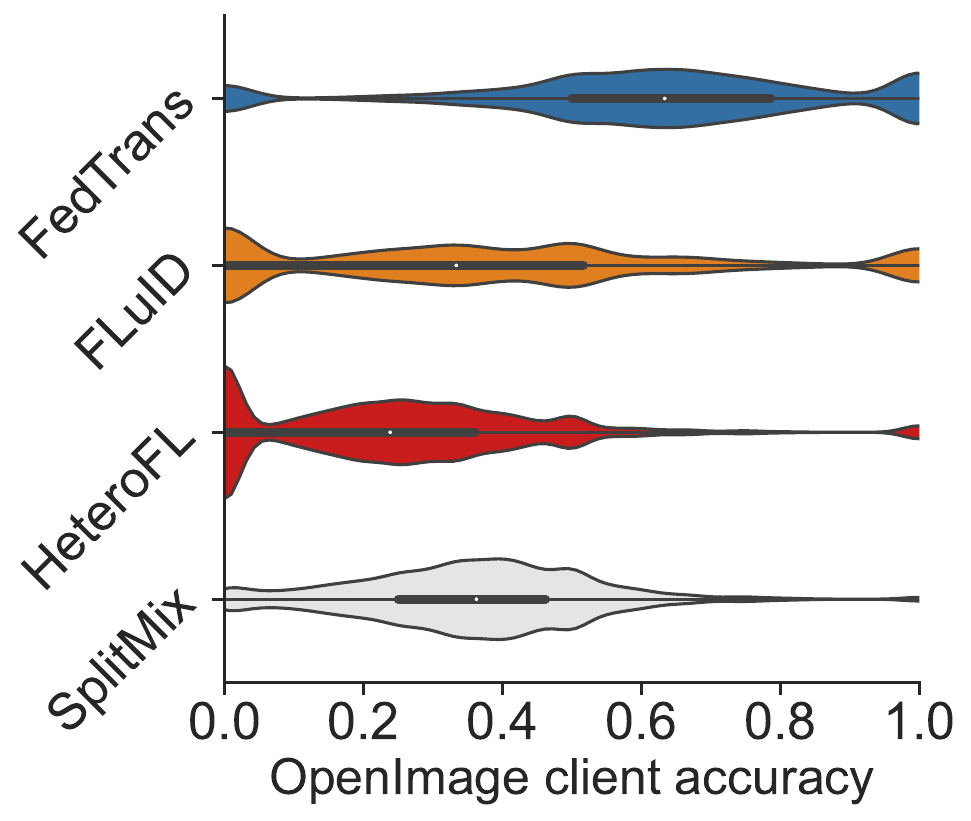}
        \captionsetup{width=0.9\linewidth}
        \label{fig:openimage}
    \end{minipage}
    \vspace{-.5cm}
    \caption{\name improves individual client model accuracy over state-of-the-art multi-model federated learning on four datasets.}
    \label{fig:state-accuracy}
\end{figure*}

\paragraph{Parameters}
We set default values for key \name parameters: layer activeness threshold, $\alpha$, is set to 0.9; the number of consecutive slopes to compute $\DoC$, $\gamma$, is 10; and the $\DoC$ threshold for $\beta$ transformation is 0.003. We show the robustness of \name benefits across these parameters in our ablation study (\S\ref{subsec:ablation}). Other FL hyperparameters are available in Appendix~\ref{subsec:expset}.

\paragraph{Baselines}
We compare \name with state-of-the-art solutions for multi-model federated learning: HeteroFL~\cite{heterofl} and SplitMix~\cite{splitmix}, and a \CR{dropout-based FL solution, FLuID~\cite{fluid}}.

\paragraph{Metrics}
We care about \emph{model accuracy} and the \emph{total training costs} to achieve them. We evaluate each client only on its compatible models and assign it the model with the highest utility. We report the average accuracy of all clients. We measure training costs using the widely used total number of MAC operations performed by all clients~\cite{energy-mac,mlsys-mac,distsys-mac}.

For each experiment, we report the mean value over 3 runs.

\subsection{End-to-End Performance}
\label{subsec:e2e}

Table~\ref{tab:perf} summarizes the key accuracy performance and training costs on all datasets.

\paragraph{\name improves model accuracy over state-of-the-art solutions.}
FedTrans improves the average model accuracy over HeteroFL by 13\% or more (Table \ref{tab:perf}). The box plots in Figure~\ref{fig:state-accuracy} zoom into the accuracy distribution of individual clients. \name clearly improves model accuracy for individual clients and achieves the best average accuracy on all four datasets. Note that HeteroFL benefits the largest model among its trained models the most, yet results in lower accuracy for smaller models. As such, HeteroFL has extremely low accuracy for clients with low hardware capacity.

\begin{figure*}[t]
    \centering
    \begin{minipage}{.24\linewidth}
        \centering
        \includegraphics[width=\linewidth]{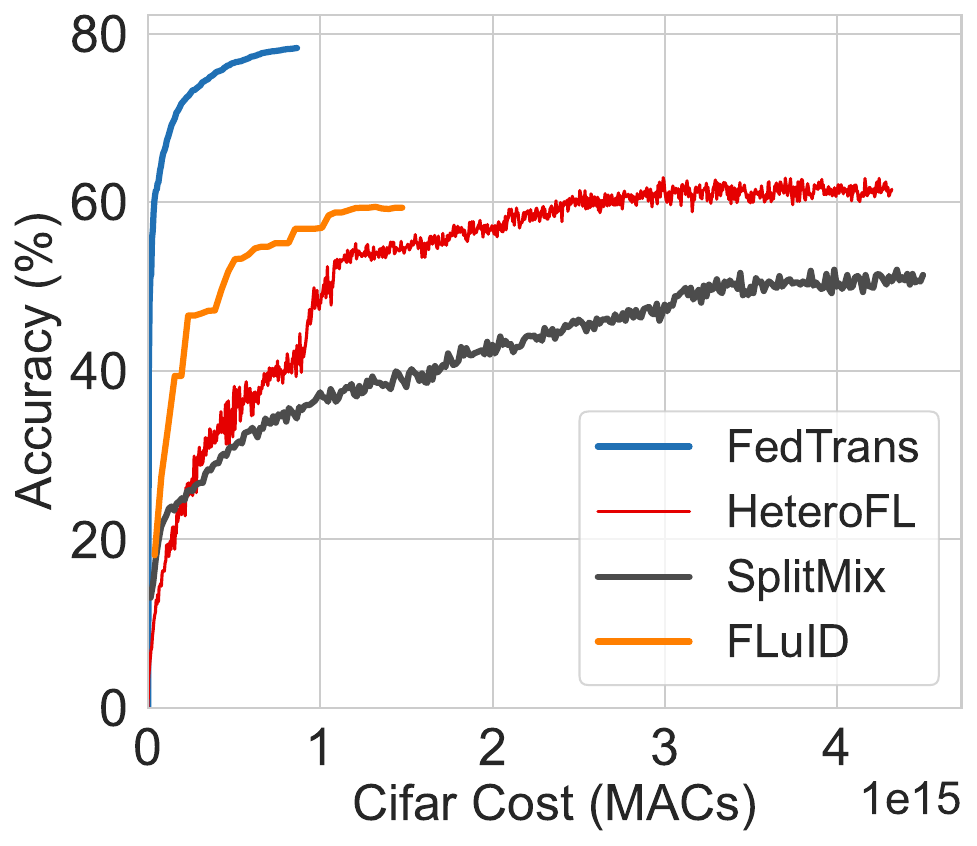}
        \captionsetup{width=0.9\linewidth}
        \label{fig:cifar-c2a}
    \end{minipage}%
    \hfill
    \begin{minipage}{.24\linewidth}
        \centering
        \includegraphics[width=\linewidth]{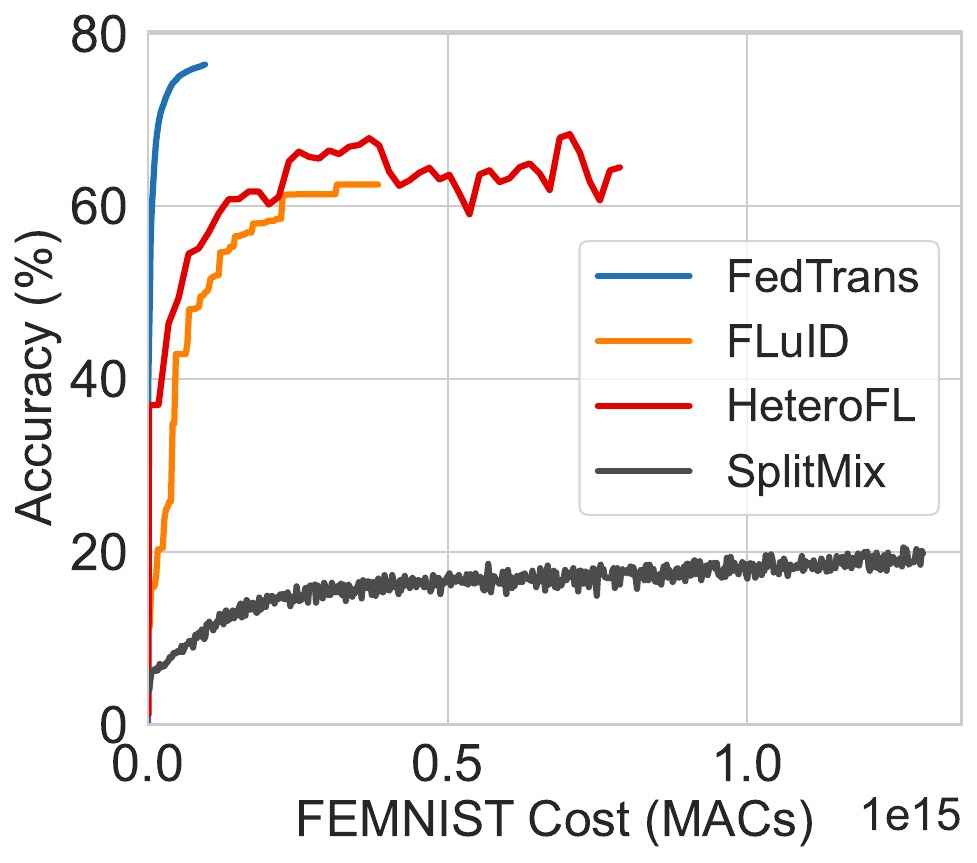}
        \captionsetup{width=0.9\linewidth}
        \label{fig:femnist-c2a}
    \end{minipage}%
    \hfill
    \begin{minipage}{.24\linewidth}
        \centering
        \includegraphics[width=\linewidth]{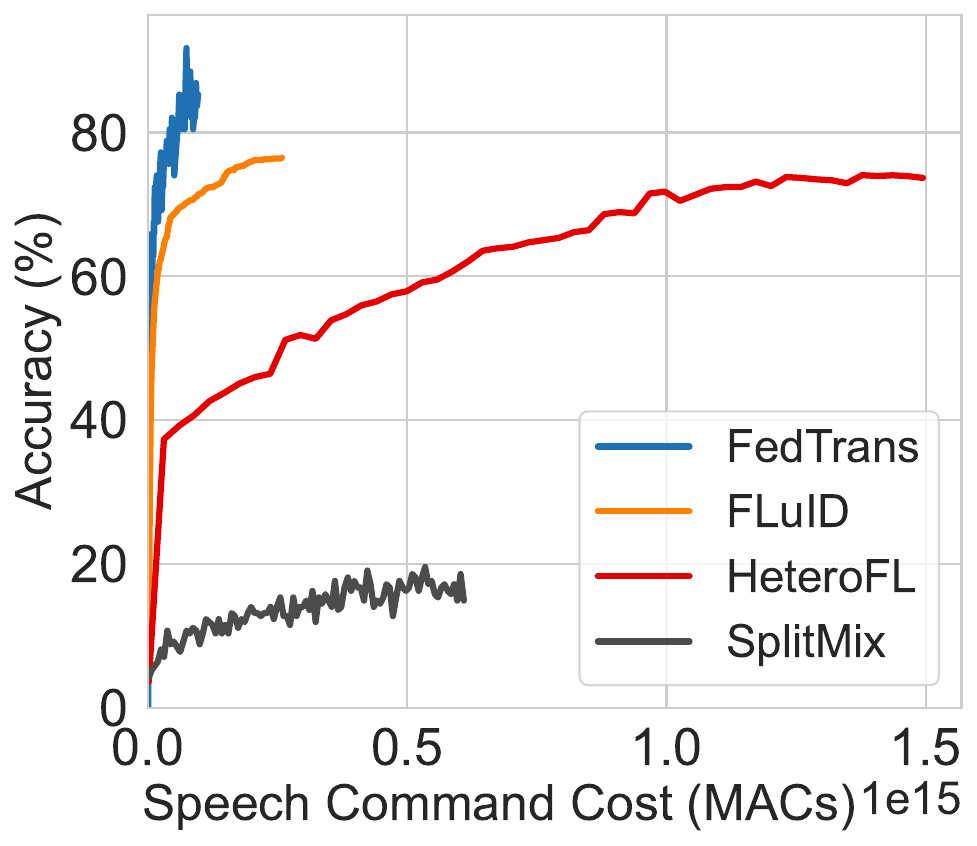}
        \captionsetup{width=0.9\linewidth}
        \label{fig:speech-c2a}
    \end{minipage}%
    \hfill
    \begin{minipage}{.24\linewidth}
        \centering
        \includegraphics[width=\linewidth]{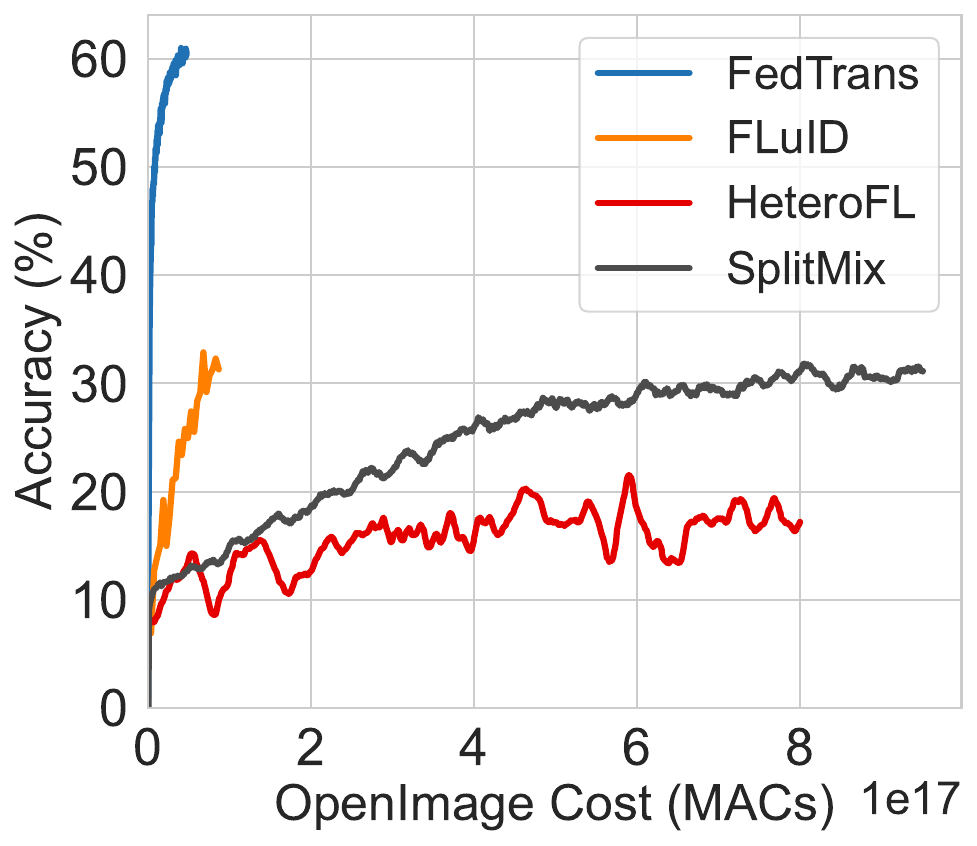}
        \captionsetup{width=0.9\linewidth}
        \label{fig:openimage-c2a}
    \end{minipage}
    \vspace{-.5cm}
    \caption{
    \name reduces training costs over state-of-the-art multi-model federated learning on four datasets.}
    \label{fig:state-cost}
\end{figure*}

\paragraph{\name reduces training costs} 
Table~\ref{tab:perf} also reports that \name reduces the total training costs by more than 4$\times$. We also notice that \name has the lowest storage footprint and the network transmission volume. Figure~\ref{fig:state-cost} further validates \name's benefits on cost-to-accuracy performance: \name incurs the lowest MAC costs toward achieving the same model accuracy. This is because \name begins with smaller models and judiciously introduces additional ones.

\begin{figure}[t]
    \begin{minipage}{.49\linewidth}
        \includegraphics[width=\linewidth]{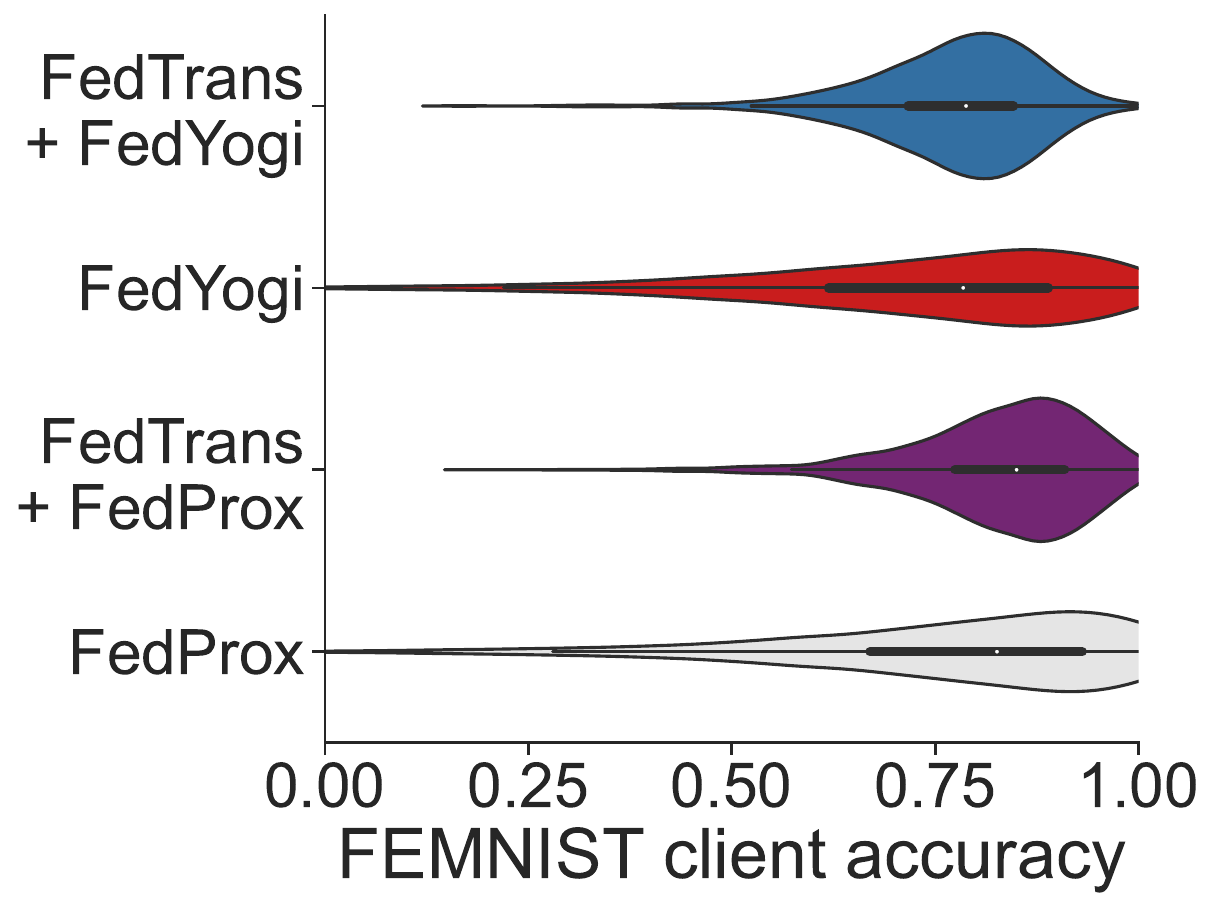}
        \caption{\name complements existing FL optimizations and improves their performance.}
        \label{fig:yogi-prox}
    \end{minipage}%
    \hfill
    \begin{minipage}{.485\linewidth}
        \centering
        \includegraphics[width=\linewidth]{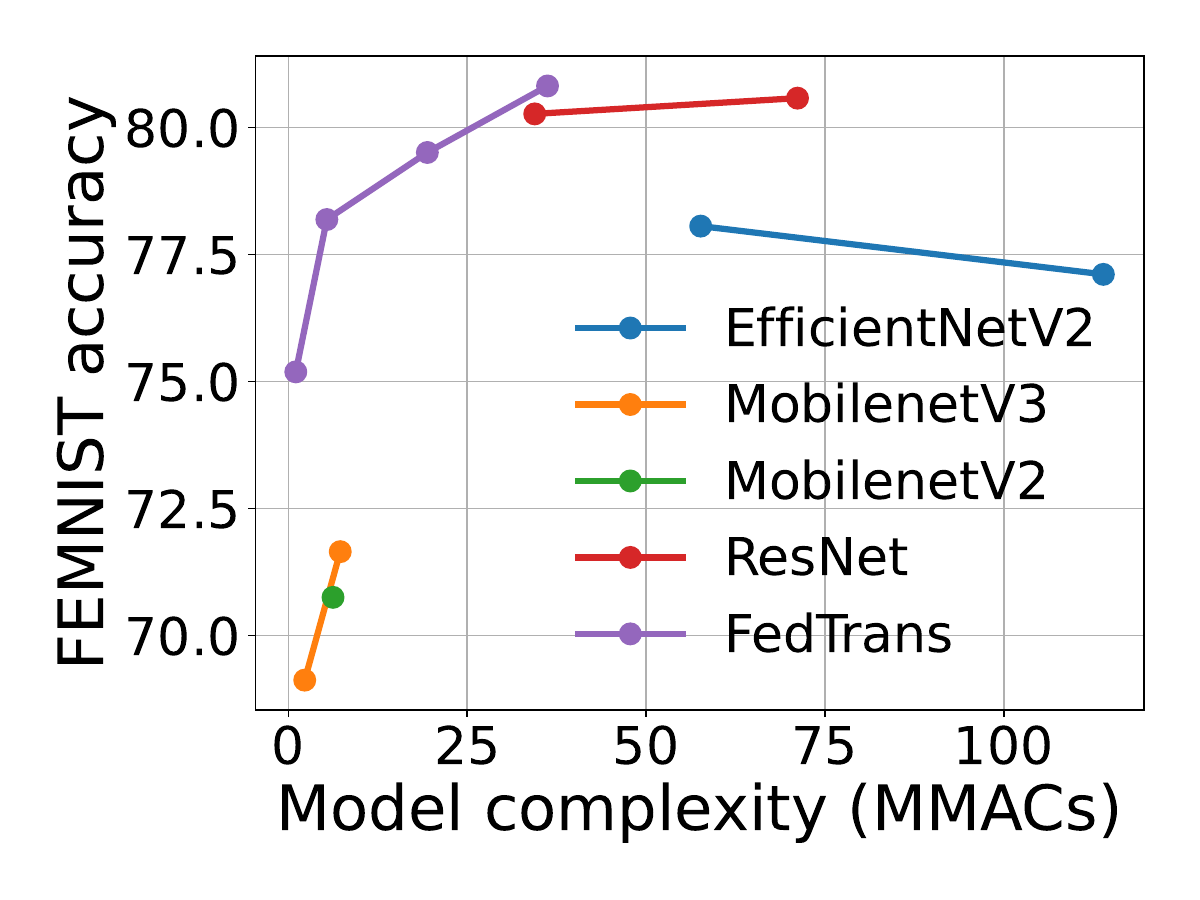}
        \caption{\name finds better model architectures for individual clients (FEMNIST dataset).}
        \label{fig:archi}
    \end{minipage}
\end{figure}

\paragraph{\name improves existing optimizations for FL training.} We implemented \name as a complementary component to the popular optimization algorithms \texttt{FedProx} and \texttt{FedYogi}. We run \name with \texttt{FedProx} or \texttt{FedYogi} on FEMNIST dataset with the NASBench201 base model as the initial model. We run \texttt{FedProx} and \texttt{FedYogi} solely with the middle-sized model generated by \name. We report the average test accuracy and training cost as the metrics. Figure \ref{fig:yogi-prox} shows that \name can improve \texttt{FedProx} and \texttt{FedYogi}, achieving higher average accuracy with the same training cost.

\paragraph{\name models outperforms state-of-the-art models for clients.}
We compare \name-generated models with state-of-the-art models on the FEMNIST dataset: EfficientNet-V2, MobileNet-V2, MobileNet-V3, ResNet-18, and ResNet-34. We sampled four of our transformed architectures from \name. Note that all \name models are transformed from the base model of NASBench201, and we measured their average accuracy. Figure \ref{fig:archi} shows that \name-generated models achieve a better tradeoff between MACs and accuracy over today's advanced models, owing to our heterogeneity-aware model assignment. 

\subsection{Performance Breakdown}
\label{subsec:breakdown}

We next analyze the impact of each component of \name on the final performance, in terms of average accuracy and training costs. We have three major components in our system: transformation-based model expansion, gradient-based layer selection, and soft aggregation across models.

As shown in Table \ref{tab:breakdown}, all components contribute to the overall performance, indicating the effectiveness and necessity of all \name components. By replacing the gradient-based layer selection with the random layer selection, the accuracy witnessed an around 3\% drop since random selection does not always pick the best layer to expand. If we further disable the weight sharing among different models, the accuracy drops by 8\%. By removing the warmup transformation and re-initializing the weights of larger models, the accuracy drops, and training cost increases by 1.6$\times$. Finally, model accuracy drops by 8\% after removing the decay factor in soft aggregation.

\begin{table}[t]
    \centering
    \begin{tabular}{l|c|c}
        \toprule
        Breakdown & Accu. (\%) & Costs (MACs) \\
        \midrule
        \name & \textbf{76.42} & \textbf{9.68} $\times 10^{14}$ \\
        \name - l & 73.44 & 9.17 $\times 10^{14}$  \\
        \name - ls & 65.16 & 9.75 $\times 10^{14}$ \\
        \name - lsw & 64.40 & 24.73 $\times 10^{14}$ \\
        \name - lswd & 55.90 & 3.37 $\times 10^{14}$ \\
        \bottomrule
    \end{tabular}
    \captionsetup{width=0.9\linewidth}
    \captionof{table}{Performance breakdown. `l' means layer selection, `s' means soft aggregation, `w' means warm up, and `d' means decayed weight sharing.}
    \label{tab:breakdown}
\end{table}

\subsection{Ablation Study}
\label{subsec:ablation}

We next analyze the impact of parameters and data heterogeneity on \name performance. \CR{We present the results of the FEMNIST dataset with similar trends observed on other datasets.}

\begin{figure}[t]
    \centering
    \begin{subfigure}{.24\textwidth}
        \centering
        \includegraphics[width=\textwidth]{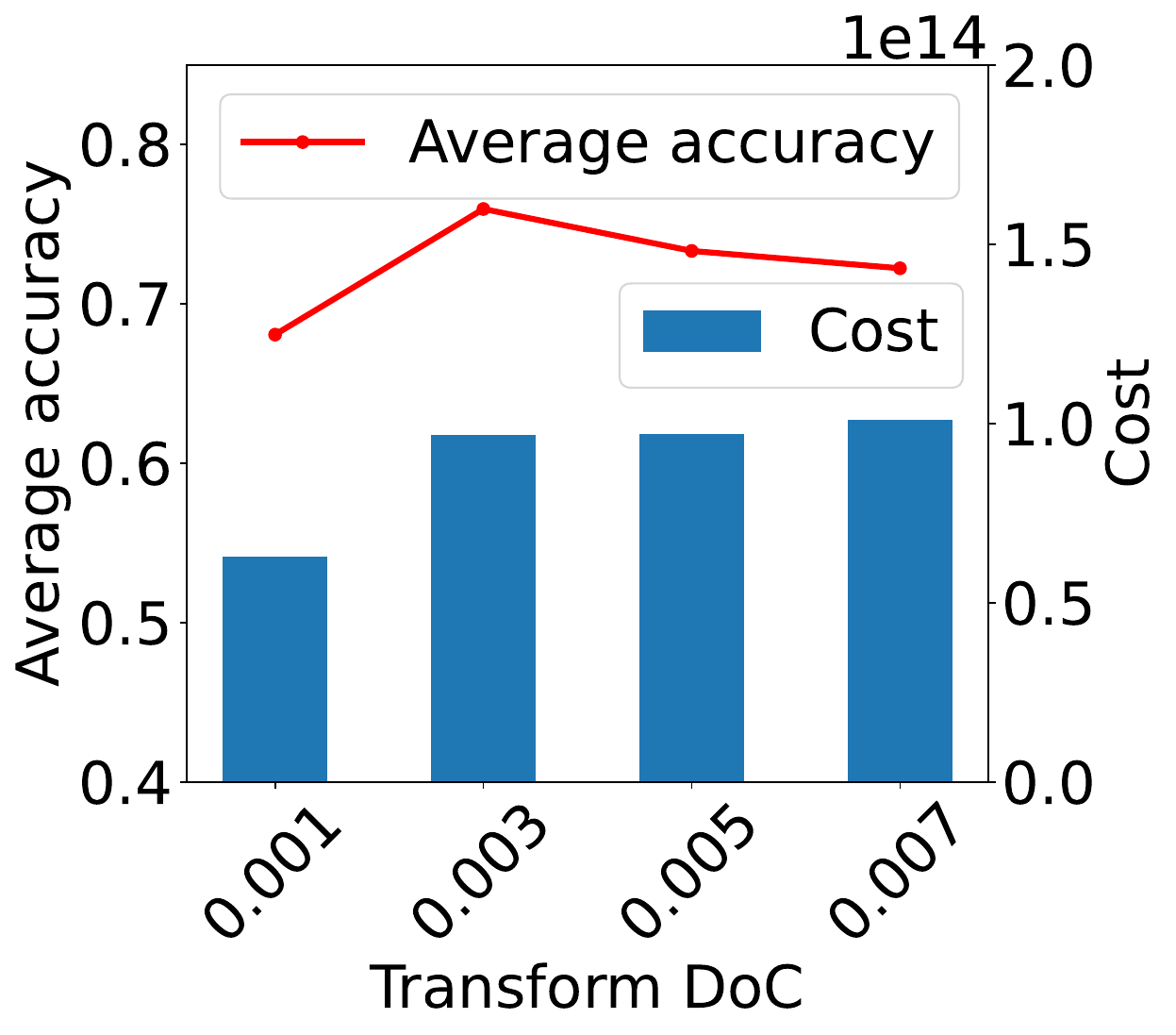}
        \subcaption{Degree of convergence ($\beta$).}
        \label{fig:doc}
    \end{subfigure}
    \hfill
    \begin{subfigure}{.235\textwidth}
        \centering
        \includegraphics[width=\textwidth]{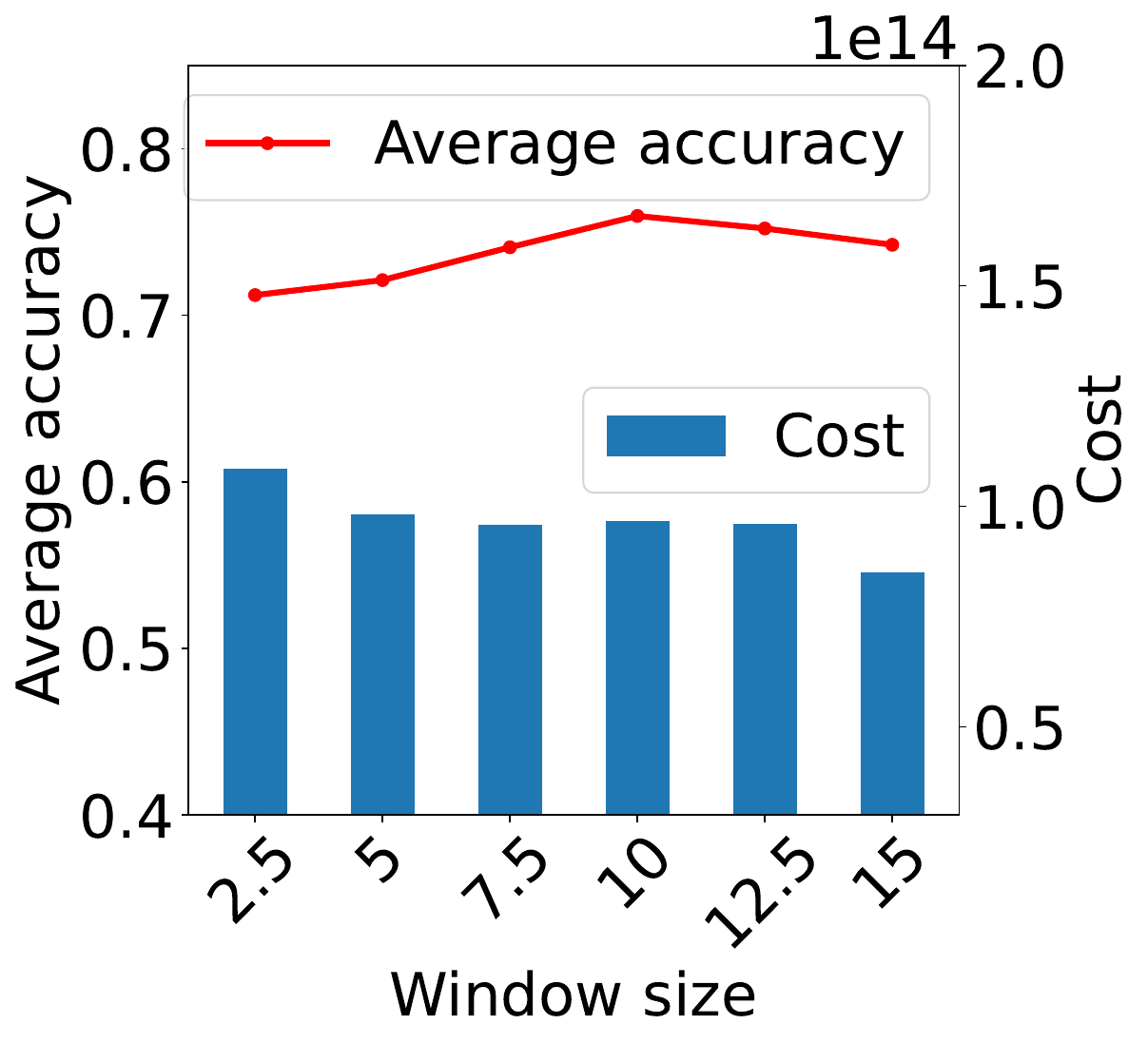}
        \subcaption{\#slops for $\DoC$ ($\gamma$).}
        \label{fig:windowM}
    \end{subfigure}
    \caption{\name achieves robust performance improvement by picking the right time to transform.}
\end{figure}

\textbf{Threshold of $\DoC$ to transform ($\beta$)}. 
Figure \ref{fig:doc} shows that as $\beta$ increases, a model is more easily considered to be ready for transformation, leading to more models transformed in the whole training process and thus higher training costs. Moreover, the accuracy is higher as $\beta$ increases initially because \name has more models capturing the data characteristics of clients comprehensively. However, the accuracy significantly drops when $\beta$ is too high because the data samples per model are too few.

\textbf{Number of consecutive slops to calculate $\DoC$ ($\gamma$)}. 
Figure \ref{fig:windowM} shows that increasing $\gamma$ generally increases the difficulty of reaching a certain DoC because we are taking more consecutive loss slopes into the average. Therefore, increasing $\gamma$ significantly decreases the training cost before the model is transformed fewer times. On the other hand, increasing $\gamma$ can help improve the final accuracy as more data samples are used to train one model, increasing the generality of the model. However, if $\gamma$ becomes too large, \name has too few models to capture the heterogeneity of clients' data.

\begin{figure}[t]
    \begin{minipage}{.48\linewidth}
        \includegraphics[width=\textwidth]{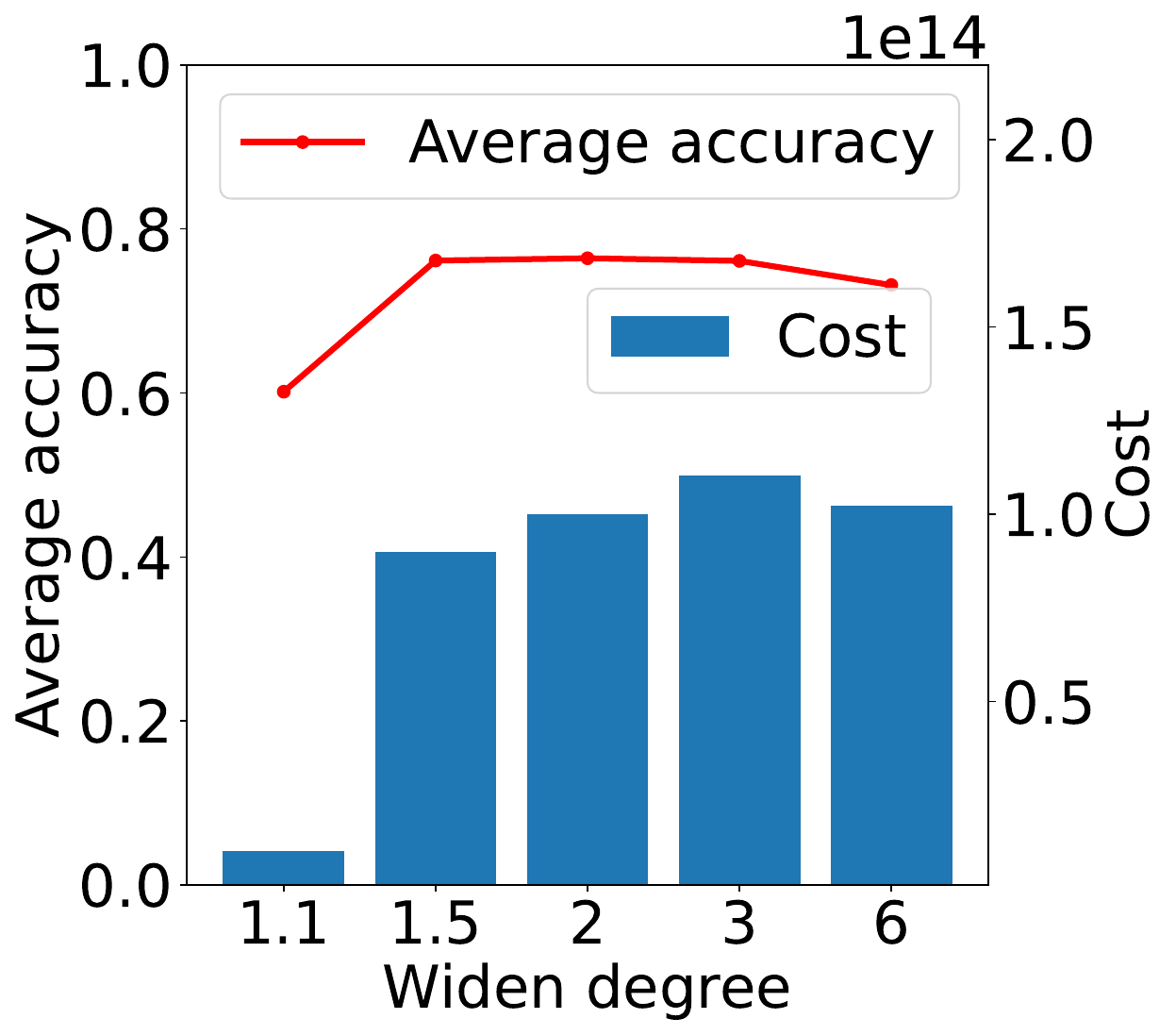}
    \end{minipage}%
    \hfill
    \begin{minipage}{.48\linewidth}
        \centering
        \includegraphics[width=\textwidth]{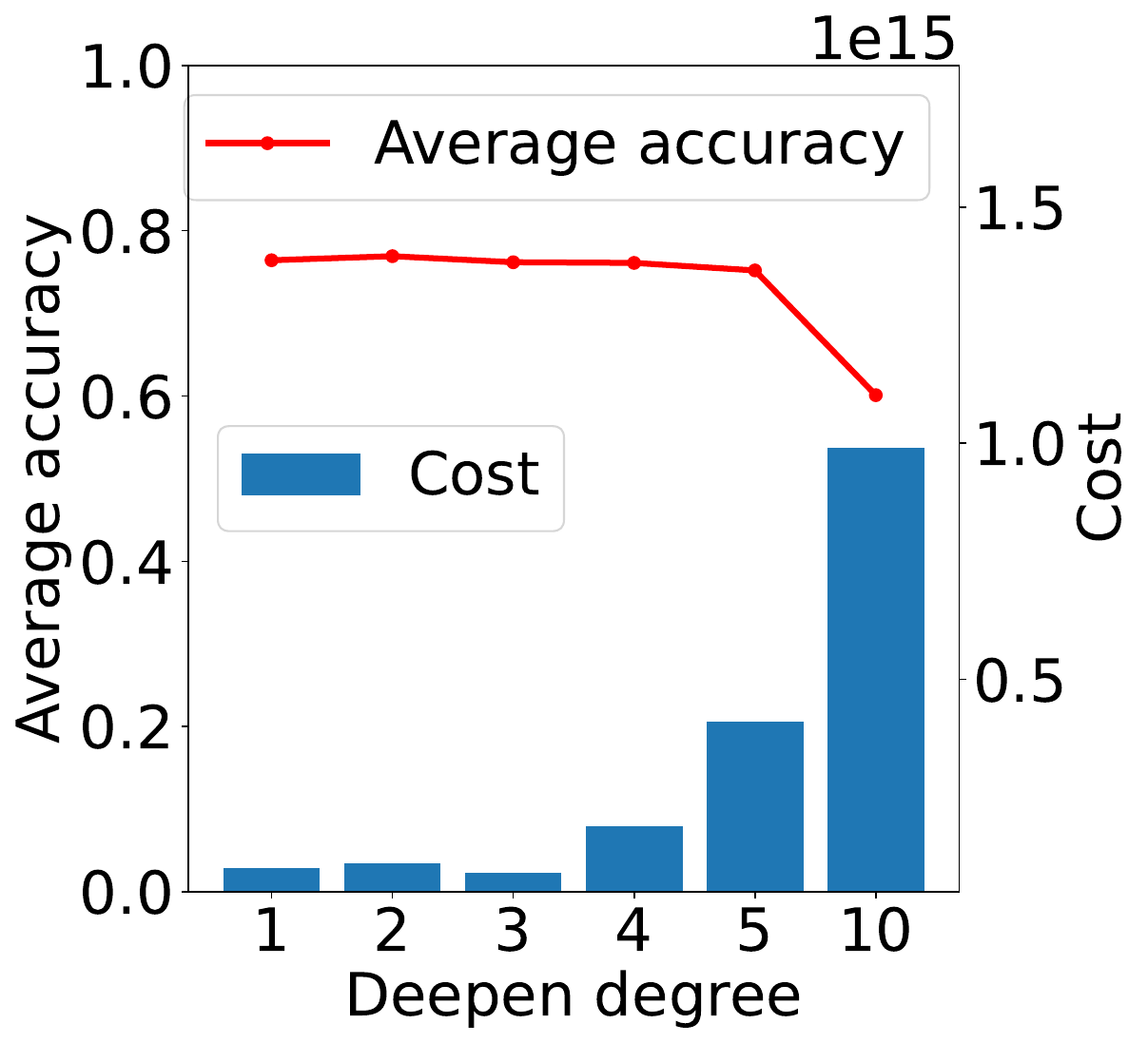}
    \end{minipage}
    \caption{\name is robust to different transformation degrees.}
    \label{fig:widen-deepen}
\end{figure}

\textbf{Impact of Widening and Deepening Degrees}
Figure \ref{fig:widen-deepen} shows that the average test accuracy and training cost are robust to a wide range of widening or deepening degrees. Intuitively, when the widening or deepening degree is slightly larger, \name trains fewer large models but earlier, which does not change the training cost. Meanwhile, although larger degrees decrease the total number of models transformed, each model is more aggressively optimized and gains more capability.

\begin{table}[htbp]
    \centering
    \begin{tabular}{ccc}
    \toprule
    Method   & Accu. (\%) & Cost (MACs) \\ 
    \midrule
    FedTrans + FedAvg & 76.5 & 3.8$\times 10^{11}$ \\
    FedAvg & 71.5 & 1.09$\times 10^{13}$ \\
    \bottomrule    
    \end{tabular}
    \caption{\name improves for ViT models (FEMNIST dataset).}
    \label{tab:vit}
\end{table}

\paragraph{\name optimizes for a wide range of models.} 
Prior work is mostly limited to convolution networks~\cite{heterofl,splitmix}, while \name is generalizable to many model architectures, including ViT~\cite{vit}. Table~\ref{tab:vit} shows that \name outperforms existing FL optimization on ViT models with a 5\% improvement of accuracy and saving costs by orders of magnitude.

\begin{figure}[t]
    \begin{minipage}{.47\linewidth}
        \includegraphics[width=\textwidth]{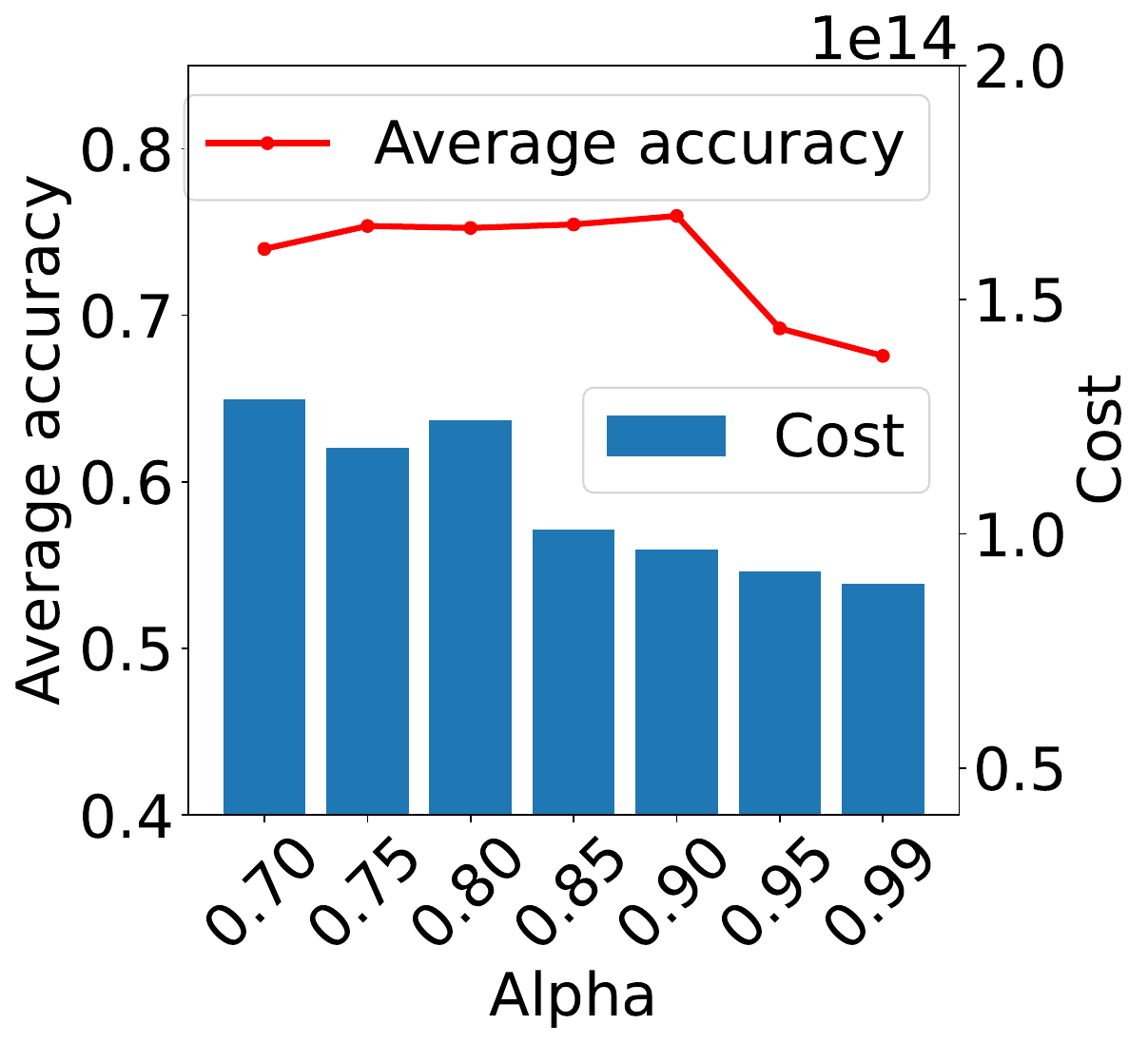}
        \caption{\name improves performance by picking the right \cell to transform.}
        \label{fig:alpha}
    \end{minipage}%
    \hfill
    \begin{minipage}{.5\linewidth}
        \centering
        \includegraphics[width=\textwidth]{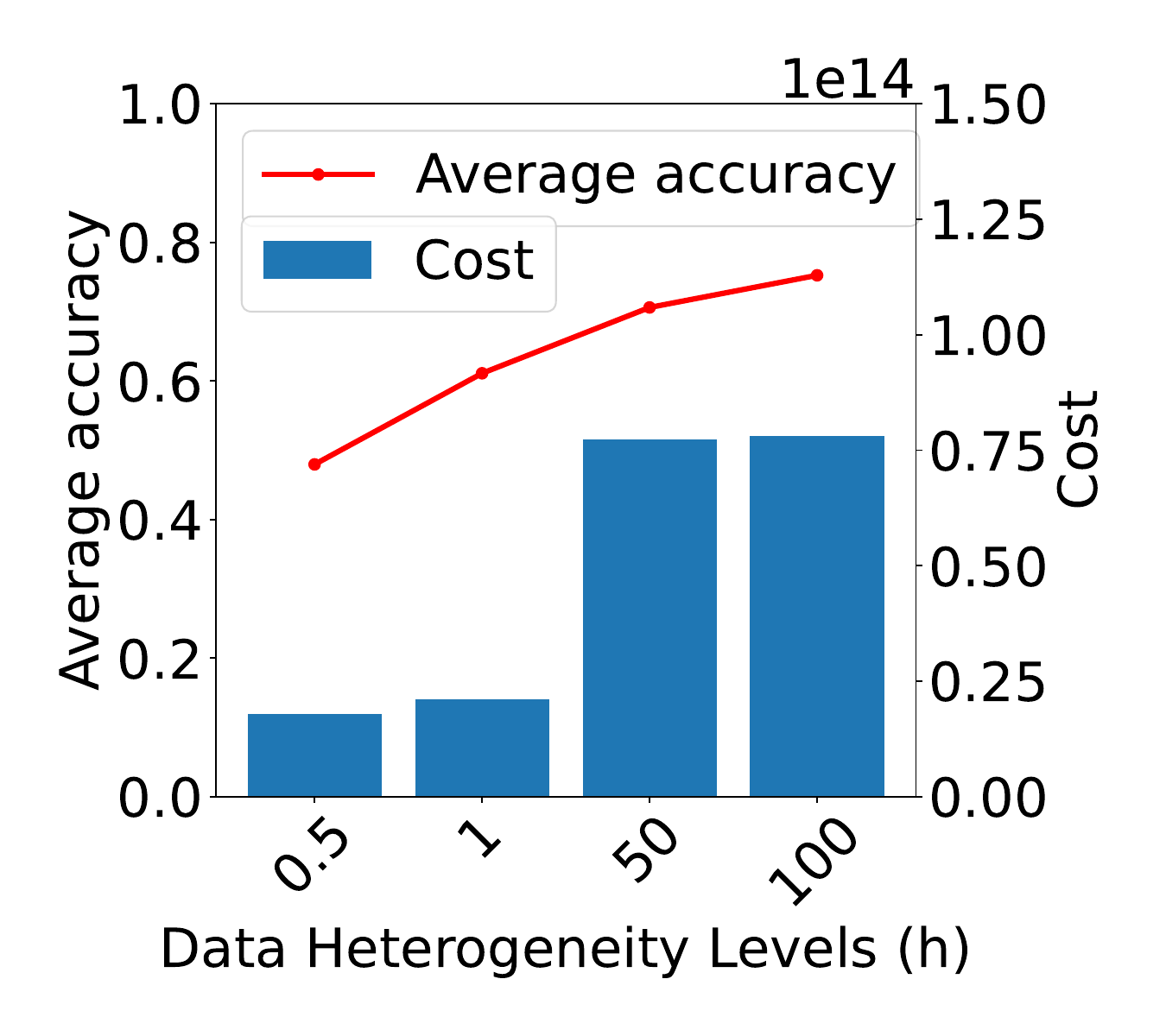}
        \caption{\CR{Performance of \name under different levels of data heterogeneities.}}
        \label{fig:hetero}
    \end{minipage}
\end{figure}

\textbf{Threshold of layer activeness to transform ($\alpha$)}. 
Figure \ref{fig:alpha} shows that as $\alpha$ increases, fewer layers are selected to expand, making the expanded model smaller and thus decreasing the training cost. On the other hand, the accuracy drops after $\alpha$ reaches 0.9, indicating that too few layers are selected to expand the capability of the model sufficiently.

\textbf{Impact of Data heterogeneity} 
Figure \ref{fig:hetero} shows the change in average accuracy and training cost while we tune the synthetic data heterogeneity of the FEMNIST dataset. Similar to the prior work \cite{heterofl, fedrolex}, we synthesize different data heterogeneity levels by controlling the label distribution with a Dirichlet distribution and parameter $h$. The lower $h$, the higher the heterogeneity.

\CR{Under low data heterogeneity, \name converges with better accuracy using more rounds. Therefore, the training costs are seemingly high (Fig. \ref{fig:hetero}). Meanwhile, the performance of \name diminishes under high data heterogeneity, which underscores the need for future research to enhance the training algorithms of multi-model FL in highly heterogeneous settings.}
\section{Related Work}
\label{sec:related}

\paragraph{System optimizations for FL training} 
Several studies propose to optimize the FL training from a system scheduling perspective. \cite{oort} prioritizes clients with good data quality and hardware capability to improve efficiency, while~\cite{pyramidfl} further investigates the data and system heterogeneity of selected clients to optimize the profiling. Besides, clustering-based optimization is proposed to mitigate the data heterogeneity by only aggregating the local models of clients with similar data distribution~\cite{auxo-socc,clustering2}, while \cite{meta_async} proposes a scalable asynchronous training scheduling algorithm to address straggler issues in synchronous FL.

\paragraph{Multi-model FL}
Training customized models for clients at scale is challenging due to the data and system heterogeneity, as well as execution costs. Weight-sharing mechanisms have been proposed to train multiple models simultaneously to save training costs, but their exploration of model architectures is either static~\cite{heterofl} or tailors only the width of the model~\cite{hermes, fjord, hetero_nips22}, resulting in suboptimal performance. Secondly, previous work on applying NAS algorithms to FL settings either uses a super-model reduction method, which introduces an expensive search phase~\cite{fednas}, or randomly samples architectures, which are usually of low quality~\cite{fedoras}. Recent work on personalized FL uses multi-task learning~\cite{multi_pfl1, multi_pfl2}, meta learning~\cite{meta_pfl1, meta_pfl2}, and other ML-based techniques~\cite{pfl1,pfl2} to optimize the accuracy performance, while neglecting the system heterogeneity.

\paragraph{Model Transformation}
Prior work proposes techniques of neural network morphism \cite{net2net,network_morphism,ofa} to transform a neural network to a larger one, preserving the complete functionality. Such techniques have been applied to reinforcement learning~\cite{mixm}, transfer learning~\cite{progvit}, and neural architecture search algorithms~\cite{nas,drnas} to save the costs of training new and larger models. Recent work in cloud computing leverages the model transformation techniques to accelerate training on the cloud without sacrificing the accuracy~\cite{modelkeeper}. \name further applies model transformation to distributed neural network training in a heterogeneous and federated setting.
\section{Conclusion}
This paper introduces \name, a novel multi-model FL training framework. \name employs \cell-wise model transformation to efficiently generate models for clients with heterogenous data and system capabilities. It uses an adaptive model assignment mechanism to explore the right model for individual clients during training, and performs inter-model aggregation to minimize the costs of training multiple models.
Our evaluations using real-world datasets show that \name significantly improves model accuracy while reducing training costs.

\section*{Acknowledgement}
\CR{We thank the anonymous reviewers, our shepherd Kevin Hsieh, and SymbioticLab members for their invaluable comments and suggestions, which greatly improved the quality of the paper. We are grateful to the CloudLab team for providing computing resources for \name experiments. Additionally, we thank Xingran Shen and Xiang Sheng for their insight in developing \name design. This work was supported in part by NSF grant CNS-2106184 and a grant from Cisco.}

\bibliography{fedtrans} 
\bibliographystyle{mlsys2024} 

\clearpage

\appendix
\section{Appendix}

\subsection{Experiment settings}
\label{subsec:expset}
\begin{figure*}[htb]
    \centering
    \includegraphics[width=\linewidth]{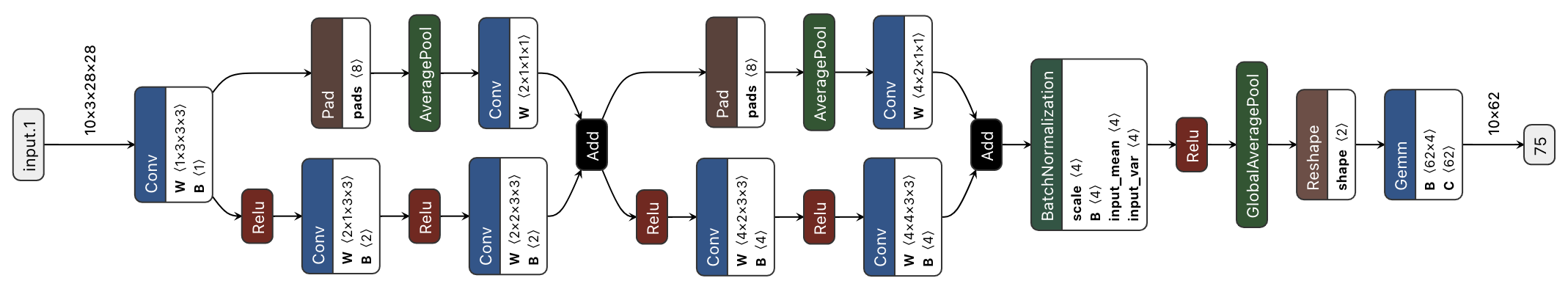}
    \caption{Base model of NASBench201}
    \label{fig:nasbench201}
\end{figure*}

\begin{figure*}[htb]
    \centering
    \includegraphics[width=\linewidth]{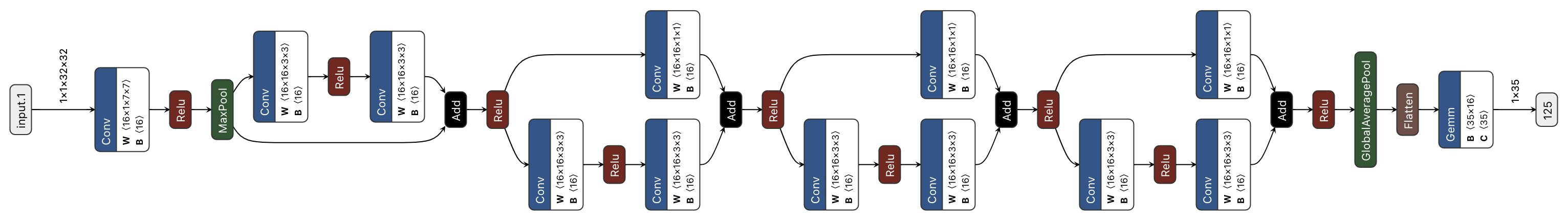}
    \caption{Modified smaller ResNet18}
    \label{fig:resnet18}
\end{figure*}

\paragraph{Comparing with baselines}
We choose MobileNetV3-small (\cite{mobilenetv3}) for Cifar-10, base model of NASBench201 (\cite{nasbench201}) for FEMNIST, and modified smaller ResNet18 (\cite{resnet}) for Speech Command and OpenImage as initial models. The detailed architecture of the base model of NasBench201 is shown in Figure \ref{fig:nasbench201}. The detailed architecture of the modified small ResNet18 is shown in Figure \ref{fig:resnet18}.

To fairly evaluate the performance across different methods, HeteroFL, SplitMix, and FLuID should use the same architecture as \name. However, HeteroFL, Splitmix, and FLuID shrink models, which means they take a large model and adopt some algorithm to reduce, compress, or prune it to form multiple small models. Therefore, we give the largest model transformed by \name as the input large model to HeteroFL, SplitMix, and FLuID. Since HeteroFL and SplitMix do not support convolutional layer with groups, we convert the grouped convolution layer to non-grouped one, which potentially increases the complexity of the layer.

The hyperparameter setting for \name is shown in Table \ref{tab:hyperparam}. The training is considered complete when either the maximum number of training rounds is reached or the validation accuracy converges, which is defined as the accuracy not improving by more than 1\% over 10 consecutive rounds. The hyperparameter settings for HeteroFL, SplitMix, and FLuID are the same as those in their paper.

\paragraph{Quality of transformed models}
To evaluate the quality of transformed models (Fig. \ref{fig:archi}), we fine-tune each transformed model on all the clients. We use the default FedAvg \cite{fedavg} setting for this evaluation part, which means we remove the hardware capacity constraints and disable the transformation, adaptive model assignment, and soft aggregation.

\section{Computation and communication overheads analysis}
\label{sec:compcomm}
Due to the challenge of data heterogeneity and the nature of distributed computing, FL training itself is expensive. Therefore, \name introduces minimal computation and communication overhead compared with standard FedAvg. 
\paragraph{Clients}
The local training on the client is the same as FedAvg, with no computation overhead. After the local training, clients are required to upload the model weights, model gradient, and training loss back to the coordinator. However, the updated model weights can be easily derived from the model gradient and the model weights of the last round. Therefore, only the training loss is considered as communication overhead for clients. Overall, on the side of clients, there is no computational overhead and negligible (\ie. a floating number) communication overhead.
\paragraph{Coordinator}
After receiving the updates from clients, the coordinator is scheduled to do four steps of computation, which are (1) updating utilities, (2) updating local weights, (3) updating the degree of convergence (DoC), and (4) model transformation. Among these steps, updating utilities, updating the degree of convergence, and model transformation are computational overhead. Given $m$ clients and $n$ models, the coordinator needs to do $m\times n$ times of utility updating operations. For each utility update, the coordinator needs to calculate the standardized loss and the subtraction, which are considered to have constant complexity. Updating DoC calculates the average of loss slopes, which is considered to have constant complexity. We consider the model transformation happens at constant times. For each model transformation, the coordinator calculates the layer activeness and applies the widening and/or deepening operations, whose complexity is considered to be proportional to the size of model weights. As for communication, \name does not introduce any overhead on the side of the coordinator.
Overall, the computational and communication overhead analysis is summarized in Table \ref{tab:overhead}.
\begin{table}[t]
    \centering
    \small
    \begin{tabular}{lc}
        \toprule
        Overhead & Estimated value \\
        \midrule
        client's computation & $0$ \\
        client's communication & $rpc$ \\
        coordinator's computation & $r(mn+1)c + |W|c$ \\
        coordinator's communication & $0$ \\
        \bottomrule
    \end{tabular}
    \captionsetup{width=0.9\linewidth}
    \captionof{table}{Computation and communication overheads analysis for $m$ registered clients, $p$ participated clients, $n$ models, $r$ rounds, where $c$ is a small constant and $|W|$ is the average size of the model weights.}
    \label{tab:overhead}
\end{table}
\begin{table}[t]
    \centering
    \small
    \begin{tabular}{ccc}
        \toprule
        Method   & Avg. (s) & Std. (s) \\ 
        \midrule
        FedTrans + FedAvg & 134.5 & 237.1 \\
        FedAvg & 226.3 & 325.6 \\
        \bottomrule    
    \end{tabular}
    \caption{Round completion time comparison.}
    \label{tab:straggler}
\end{table}

\begin{table*}[t!]
    \centering
    \footnotesize
    \begin{tabular}{lcccc}
        \toprule
        Hyperparameters & Cifar-10 & FEMNIST & Speech Command & OpenImage\\
        \midrule
        \# of participants per round & 10 & 100 & 100 & 100\\
        maximum number of training rounds    & 1000   & 2000 & 1500 & 2000\\
        step size to calculate the loss slope ($\delta$) & 20 & 30 & 100 & 50\\
        local training steps & & 20 & &\\
        batch size & & 10 & &\\
        learning rate & & 0.05 & &\\
        decay factor & & 0.98 & &\\
        \# of consecutive gradient to calculate activeness ($T$) & & 5 & &\\
        \bottomrule
    \end{tabular}
    \captionsetup{width=0.9\linewidth}
    \captionof{table}{Hyperparameters}
    \label{tab:hyperparam}
\end{table*}
\section{\name mitigates the straggler issue.} 
In synchronous federated learning, slow clients could slow down the training process if clients are given the same workload, which is referred to as the straggler issue. FedTrans can mitigate the straggler issue as we assume each client has a hard requirement for the model complexity (MACs). As shown in Table \ref{tab:straggler}, FedTrans improves FedAvg both in the average and the std of the round completion time among clients on FEMNIST dataset compared with FedAvg. 
\end{document}